\documentclass{article}

\usepackage{microtype}
\usepackage{graphicx}
\usepackage{subcaption}
\usepackage{booktabs} 
\usepackage{multirow}
\usepackage{makecell}
\usepackage{colortbl} 

\usepackage{hyperref}

\usepackage{algorithm}
\usepackage{algorithmic}

\usepackage{pifont}

\usepackage{enumitem}

\usepackage{listings}
\usepackage[most]{tcolorbox}

\PassOptionsToPackage{table,xcdraw,dvipsnames}{xcolor}
\usepackage[table,xcdraw,dvipsnames]{xcolor}
\definecolor{ForestGreen}{RGB}{34,139,34}
\definecolor{myyellow}{RGB}{181, 181, 27}
\definecolor{mygrey}{gray}{0.4}

\usepackage[preprint]{icml2026}

\usepackage{amsmath}
\usepackage{amssymb}
\usepackage{mathtools}
\usepackage{amsthm}
\usepackage{bbm} 

\usepackage[capitalize,noabbrev]{cleveref}

\theoremstyle{plain}

\theoremstyle{definition}

\theoremstyle{remark}

\usepackage[disable,textsize=tiny]{todonotes}

\newcommand{\our}{SKPO}

\newcommand{\keyword}[1]{\texttt{\textbf{#1}}}

\makeatletter
\def\Xhline#1{%
  \noalign{\ifnum0=`}\fi
  \hrule height #1
  \futurelet\reserved@a\@xhline
}
\makeatother

\begin{document}

\twocolumn[
\icmltitle{Skip-Connected Policy Optimization for Implicit Advantage}



\icmlsetsymbol{equal}{*}
\begin{icmlauthorlist}
\icmlauthor{Fengwei Teng}{equal,inst1}
\icmlauthor{Jinyi Bai}{equal,inst2}
\icmlauthor{Xinhao Yao}{inst2}
\icmlauthor{Demi Ruohan Wang}{inst3}
\icmlauthor{Jiahao Zhao}{inst2}
\icmlauthor{Zhijiang Guo}{inst4}
\end{icmlauthorlist}

\icmlaffiliation{inst1}{Mohamed bin Zayed University of Artificial Intelligence}
\icmlaffiliation{inst2}{Renmin University of China}
\icmlaffiliation{inst3}{Carnegie Mellon University}
\icmlaffiliation{inst4}{Hong Kong University of Science and Technology (Guangzhou)}

\icmlkeywords{Reinforcement Learning, Large Language Models, Policy Optimization, Reasoning, RLHF}

\vskip 0.3in
]



\printAffiliationsAndNotice{\icmlEqualContribution}

\begin{abstract}

Group Relative Policy Optimization (GRPO) has proven effective in RLVR by using outcome-based rewards. While fine-grained dense rewards can theoretically improve performance, we reveal that under practical sampling budgets, Monte Carlo estimation yields high-variance and sign-inconsistent advantages for early reasoning tokens, paradoxically underperforming outcome-only GRPO. We propose Skip-Connected Optimization (\our{}), which decomposes reasoning into upstream and downstream phases: upstream receives dense rewards from downstream Monte Carlo sampling with single-stream optimization; downstream maintains group-relative optimization, where a skip connection concatenates the upstream segment with the original problem, enabling the model to leverage helpful upstream reasoning while preserving the freedom to bypass flawed reasoning through direct problem access. Experiments demonstrate improvements of 3.91\% and 6.17\% relative gains over the strongest baselines on Qwen2.5-Math-7B and Llama-3.2-3B respectively across mathematical benchmarks and out-of-domain tasks including general reasoning and code generation. Further analysis reveals an implicit advantage: \our{} generates trajectories with higher intermediate-step quality even when matched for final correctness.


\end{abstract}

\section{Introduction}
\label{sec:introduction}

\begin{figure}[t]
    \centering
    \includegraphics[width=\linewidth]{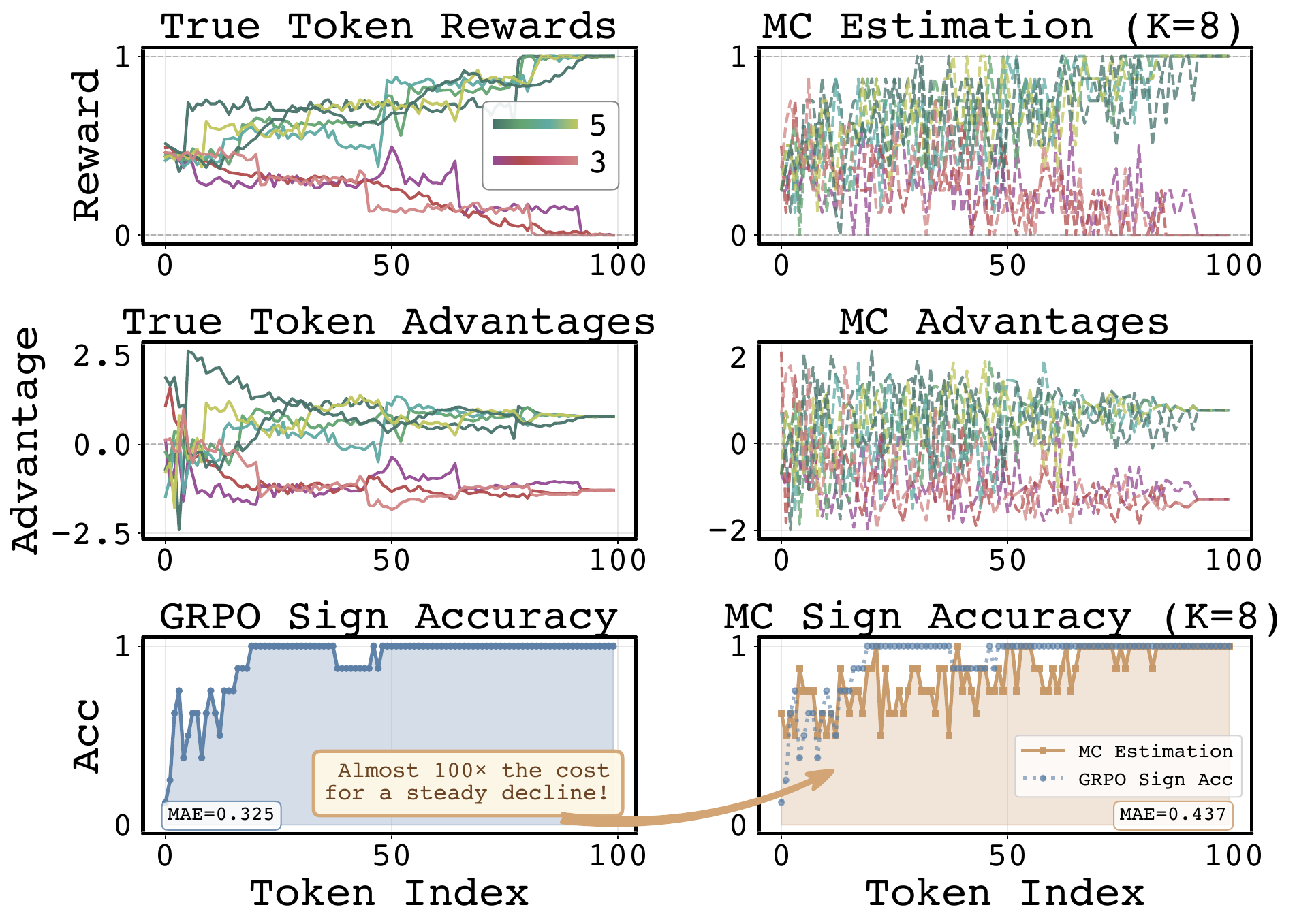}
    \caption{Monte Carlo estimation of token-level rewards under practical sampling budgets ($K=8$). Top: true values (left) vs.\ estimates (right). Bottom: sign accuracy and MAE, showing that early tokens suffer from higher estimation error. By the nature of MC sampling, true token rewards converge to the policy's average accuracy on the prompt at early positions, and to the trajectory's binary correctness at late positions; the former makes accurate advantage estimation difficult at early steps, while the latter explains why GRPO achieves reasonable posterior estimation. Additional examples across more correctness distributions (0/8 to 8/8) are provided in \Cref{app:correctness_distribution}.}
    \label{fig:trajectory_convergence}
\end{figure}

Reinforcement Learning with Verifiable Rewards (RLVR)~\cite{shao2024deepseekmath, hu2025reinforcepp, rloo2024, li2023remax} has emerged as the dominant paradigm for enhancing LLM reasoning. Among these methods, the representative Group Relative Policy Optimization (GRPO) achieves strong training results through outcome-based rewards and within-group relative advantage computation with moderate sampling budgets. While this coarse-grained binary design demonstrates stability, fine-grained dense rewards and advantages remain theoretically appealing for their higher performance upper bounds; yet limited inference budgets pose a practical challenge: \textbf{how to make stable progress toward the upper bound promised by dense fine-grained rewards and advantages, under the existing RLVR paradigm}.

Our preliminary analysis reveals a counterintuitive finding that partially explains this dilemma. By employing Monte Carlo sampling to estimate step-level rewards through sampling multiple continuations from existing reasoning and aggregating their outcome rewards~\cite{yue2025vsrm}, we obtain the results shown in \Cref{fig:trajectory_convergence}. Even with substantial sampling costs, MC estimation combined with group-relative advantage computation yields high-variance advantages with frequent sign errors for early reasoning tokens, stably underperforming outcome-only GRPO. This finding may explain why existing methods must make various trade-offs between fine-grained and coarse-grained, dense and sparse, to circumvent this estimation challenge.

Existing methods make different sacrifices in their designs. \ding{182}~\textsl{\texttt{Multi-stage methods}} represented by reflection and ensembling~\cite{gandhi2025cognitive} sacrifice fine-grained information: while they structure reasoning into explicit steps, step quality still reduces to outcome correctness, i.e., whether reflection corrects to the right answer or the ensemble converges to the correct solution. \ding{183}~\textsl{\texttt{Fine-grained reward methods}} face a similar dilemma. Algorithm-based approaches typically either aggregate token rewards within responses to compute group-relative advantages and subsequently broadcast these to every token (which is equivalent to dense outcome-level methods), or entirely sacrifice group-relative estimation in favor of algorithms like PPO. Monte Carlo estimation, while more interpretable and offering higher upper bounds for fine-grained refinement, requires prohibitive sampling for reliable advantage estimation. \ding{228}~In summary, existing approaches face a structural dilemma between \textbf{fine-grained validity} and \textbf{computational efficiency}.

To break this dilemma, we propose \keyword{SK}ip-connected \keyword{P}olicy \keyword{O}ptimization (\our{}), which decomposes reasoning into upstream and downstream phases. The key insight is that by instantiating Monte Carlo sampling at a \textit{single} intermediate position, we can obtain dense rewards for upstream while reusing the same samples for group-relative optimization in downstream. Specifically: \ding{182}~The upstream phase generates early-stopped reasoning segments and receives Monte Carlo rewards aggregated from downstream continuations, using single-stream optimization~\cite{xu2025single} to reduce instability on fine-grained signals by lowering upstream per-prompt sample count and estimating more stable baselines from historical samples. \ding{183}~The downstream phase maintains group-relative optimization with a \keyword{skip} \keyword{connection} $[s, q]$, concatenating both the upstream segment and the original problem to balance exploration freedom and upstream influence. \ding{184}~An engineering optimization achieves computational parity with GRPO through single-pass rollout. Interestingly, this architecture also yields an \keyword{implicit} \keyword{advantage}: models trained with \our{} generate trajectories with significantly higher intermediate-step quality even when matched for final outcome correctness. This benefit emerges from the fine-grained optimization structure without external supervision.

Our contributions are:
\begin{itemize}[leftmargin=*]
    \item We identify why fine-grained Monte Carlo rewards fail under practical budgets: high-variance advantages whose signs frequently flip, underperforming outcome-only GRPO.
    \item We propose \our{}, decomposing reasoning into upstream and downstream phases, instantiating Monte Carlo sampling at a single intermediate position to provide dense rewards for upstream, while using a skip connection $[s, q]$ that enables downstream to leverage upstream reasoning while preserving exploration freedom for group-relative optimization.
    \item Experiments show substantial improvements on math benchmarks with robust OOD generalization. Analysis reveals an \keyword{implicit} \keyword{advantage}: higher intermediate-step quality emergent from the optimization structure.
\end{itemize}

\section{Related Work}
\label{sec:related}

\subsection{Group-Based Policy Optimization}
Group Relative Policy Optimization (GRPO) has emerged as a dominant paradigm for reinforcement learning from verifiable rewards~\citep{shao2024deepseekmath, deepseekr1, yu2025dapo, zheng2025gspo, wu2025grpodpo, yao2025offpolicy, jain2025landscape}. By sampling multiple responses within a group and computing relative advantages, GRPO eliminates the need for explicit value models while effectively guiding policy updates. Recent research has systematically explored and optimized various components within GRPO~\citep{khatri2025scalerl, he2025justrl}. 

Variants of GRPO explore different combinations of design choices within the framework. \ding{182}~\textit{Reward and advantage computation} encompasses methods that improve sample efficiency through decoupled reward modeling and modified advantage computation, while also modeling the influence of entropy mechanisms and step-level fine-grained information on training dynamics~\citep{rloo2024, drgrpo2025, aapo2025, chen2025drpo, jain2025landscape, he2025responselevel, he2025unlikely, chen2025dra}. \ding{183}~\textit{Policy ratio design and aggregation} includes clipping mechanisms that enhance training stability through dynamic token-level bounds and gradient-preserving designs~\citep{yu2025dapo, yang2025dcpo}, as well as normalization strategies that stabilize policy updates through ratio normalization and geometric mean aggregation, or reduce sensitivity to outliers through group expectation sampling~\citep{gmpo2025, zhang2025gepo, sheng2025espo}. \ding{184}~\textit{Alternative baseline estimation} addresses the limitation that group-relative baselines require multiple concurrent samples per prompt. SPO~\citep{xu2025single} replaces the \textit{spatial} baseline (group mean from concurrent rollouts) with a \textit{temporal} baseline via per-prompt value trackers, enabling optimization when only a single sample is available, which motivates our adoption of temporal baselines for upstream optimization where only a single segment is generated per prompt. Among these design choices, we are particularly interested in the construction of step-level information.

\subsection{Fine-Grained Reward/Advantage}

A well-known characteristic of vanilla GRPO is its uniform credit assignment: the same advantage is distributed equally across all tokens in a response, which may lead to length bias and inefficiency~\citep{liu2025dler}. Recent work addresses this limitation through two primary paradigms. \ding{182}~\textit{Algorithm-based methods} derive dense reward signals from existing rollouts through algorithmic formulations rather than additional sampling, including reformulating the Bellman equation to decompose action-level Q-values into token-level components via soft updates, and synthesizing process rewards from policy log-ratios, entropy weights, and learnable token preferences~\citep{wen2025etpo, chen2025discpo, cui2025prime, tan2025gtpo, xie2025capo, deng2025thr, li2025red, liu2025dler, lee2025sgrpo}; other works infer implicit process rewards directly from policy rollouts without explicit reward model training~\citep{fan2025posterior, fei2025spro, sullivan2025grpo, prakash2025zero, ma2025stepreward}; additionally, some methods selectively optimize only high-entropy tokens or identify pivotal reasoning steps for targeted credit assignment~\citep{wang2025forkingtokens, wang2025hicra, gpo2025}. \ding{183}~\textit{Sampling-based methods} estimate segment-level or token-level values through additional Monte Carlo sampling at intermediate states, utilizing vine sampling at segment boundaries, tree-based trajectory organization, or repeated rollouts to obtain unbiased value estimates~\citep{kazemnejad2025vine, brantley2025oar, lin2025tepo}. Algorithm-based approaches achieve fine-grained credit assignment with minimal overhead, while sampling-based methods provide accurate estimation at a computational cost proportional to the number of estimation points. However, the instability introduced by fine-grained information may outweigh its benefits compared to coarse-grained but stable optimization over long training processes.

\subsection{Multi-Stage Pipeline Designs}
GRPO-based methods also integrate with pipeline designs to enhance training effectiveness~\citep{yan2025luffy, chen2025minimax}. Beyond single-response optimization discussed above, multi-stage pipelines involve multiple responses or attempts to solve a task. \ding{182}~\textit{Structured decomposition} exploits multi-response generation to improve credit assignment, organizing multiple rollouts into search trees to track retrospective credit and identify pivotal nodes, assigning credit to high-level thought templates across hierarchical reasoning traces, or extending Monte Carlo sampling with tree-based trajectory organization for unbiased value estimation over multiple attempts~\citep{zhang2025criticsearch, yang2025reasonflux, kazemnejad2025vine, brantley2025oar}. \ding{183}~\textit{Internal feedback} constructs multi-turn refinement loops where models learn from their own previous outputs, employing multi-stage RL training with reward shaping to prevent behavior collapse, using layered generation processes where later layers explicitly correct earlier outputs under implicit process rewards, or introducing verify-then-revise mechanisms that alternate between policy and generative verifier roles~\citep{kumar2024training, ding2025multi, jiang2025pag, ekbote2025murphy, li2025scrpo}. \ding{184}~\textit{External signals} guide multi-stage reasoning through curriculum or data generation, utilizing challenger-solver co-evolution patterns that generate questions at the solver's capability frontier, formulating curriculum selection as non-stationary multi-armed bandits for adaptive difficulty scheduling, updating model parameters during inference using pseudo-labels from majority voting, or enabling self-play through questioner-responder-verifier role alternation~\citep{rzero2025, chen2025self, zuo2025ttrl, yang2025spell}.

Our work differs from these approaches in that, while adopting an upstream-downstream two-stage generation pipeline, the actual inference cost does not exceed a single rollout, making it directly comparable to on-policy single-rollout methods rather than multi-stage approaches, while similarly introducing no external information to guide the policy model. Meanwhile, we still adopt a sampling-based approach to obtain fine-grained dense rewards, but by instantiating upstream-downstream sampling at a single position and aligning with the group sampling count, we incur no additional cost while achieving both efficient sampling and stage-wise information density.

\section{Methodology}
\label{sec:method}

\begin{figure*}[htbp]
    \centering
    \includegraphics[width=1.0\linewidth, height=0.9\textheight, keepaspectratio]{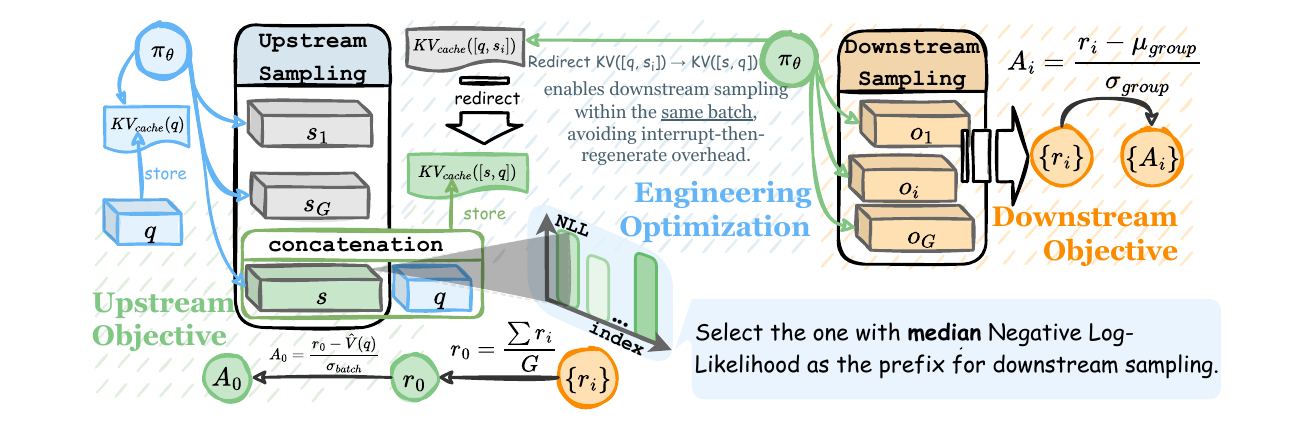}
    \caption{Overview of the \our{} algorithm. We illustrate three components: (1) upstream sampling process and components for upstream advantage computation; (2) downstream sampling process and components for downstream advantage computation; (3) engineering optimization that completes both upstream and downstream sampling within a single GPU batch through KV cache rewriting.}
    \label{fig:algo}
\end{figure*}

Figure~\ref{fig:algo} illustrates the overall algorithm. This section formalizes the key components of \our{}: \Cref{sec:mc_sampling} first analyzes ideal token-level Monte Carlo reward estimation, explaining why early-position estimation suffers from high variance while late-position estimation naturally converges to outcome rewards, a property that justifies GRPO's conservative broadcasting as a reasonable baseline. We then present the upstream-downstream architecture that instantiates Monte Carlo sampling at a single position, detailing the dual functions of the skip connection $[s, q]$, and describe an engineering optimization that achieves computational parity with GRPO. \Cref{sec:optimization} formalizes the asymmetric optimization strategy with explicit policy gradient objectives for both phases.

\subsection{Monte Carlo Trajectory Sampling}\label{sec:mc_sampling}

\paragraph{Ideal Token-Level Reward Estimation for Analysis.}
To provide fine-grained rewards for reasoning step quality, we consider token-level Monte Carlo reward estimation through trajectory sampling. For any token position $t$ in a reasoning trajectory $o_{1:T}$, we estimate dense token-level rewards by sampling $K$ continuations from the prefix $o_{1:t}$ and averaging their outcome rewards:
\begin{equation}
\label{eq:mc_token_reward}
\begin{aligned}
\mathcal{R}_{\mathrm{MC}}(o_t, a) &= \frac{1}{K} \sum_{k=1}^{K} \mathcal{R}_{\mathrm{score}}(o^{(k)}, a), \\
&\text{where } o^{(k)} \sim \pi_{\boldsymbol{\theta}}(\cdot | o_{1:t}),
\end{aligned}
\end{equation}
and $\mathcal{R}_{\mathrm{score}}(o, a)$ is a simple scoring function based on outcome correctness. In this idealized formulation, each token receives a distinct reward signal reflecting how well its prefix facilitates correct completions. Notably, as $t$ approaches the trajectory end $T$, the Monte Carlo estimates naturally converge toward the outcome reward $\mathcal{R}_{\mathrm{score}}(o, a)$, since diminishing remaining uncertainty leaves little room for divergent continuations. This convergence property places outcome-based methods such as GRPO~\cite{shao2024deepseekmath}, which directly broadcast the outcome reward to all tokens without any intermediate sampling, as the limiting behavior that Monte Carlo estimation approaches for late-trajectory tokens. While theoretically appealing for offering dense rewards at every reasoning step, this ideal token-level Monte Carlo approach faces severe practical challenges when combined with group-relative advantage computation. Computing group-relative advantages at intermediate positions requires forming stable comparison groups through extensive continuation sampling at each token location. For early reasoning tokens, accurately estimating advantage signs demands prohibitively large $K$ values, and insufficient sampling causes token-level rewards to paradoxically underperform outcome-level baselines. Conversely, GRPO's broadcasting of outcome rewards, while sacrificing fine-grained discrimination, provides a conservative estimate that avoids the catastrophic sign noise of undersampled Monte Carlo. This perspective also explains why curriculum learning~\cite{xu2025single} and resampling~\cite{yu2025dapo} improve stability: prompts with intermediate accuracy exhibit smaller reward and advantage estimation errors (see \Cref{app:correctness_distribution}).

\paragraph{Upstream-Downstream Instantiation: A Cost-Effective Step Beyond.}
To balance the computational feasibility with fine-grained rewards, we instantiate Monte Carlo trajectory sampling at a single intermediate position through an upstream-downstream architecture. The policy model $\pi_{\boldsymbol{\theta}}$ first generates an early-stopped reasoning segment $s \in \mathcal{S}$ in the upstream phase, with the segmentation position determined adaptively based on cumulative token-level statistics rather than fixed formatting rules. Specifically, for each prompt $q$, we uniformly sample a termination position $t_q \sim \mathcal{U}[\frac{1}{6}L_q, \frac{1}{2}L_q]$, where $L_q$ is the per-prompt average response length tracked via a similar KL-adaptive mechanism as the upstream value tracker (see \Cref{sec:optimization}), without warm-up to ensure cost parity with baselines. For responses that terminate before $t_q$, we use half of the actual response length as the split position. We then sample $G$ continuations from the augmented input $[s, q]$, the concatenation of the segment and original problem, to obtain complete responses $\{o_i\}_{i=1}^G$ in the downstream phase. This skip connection $[s, q]$ serves dual functions: \ding{182}~\textit{it preserves the upstream segment's influence on downstream reasoning, enabling valid Monte Carlo estimation}; \ding{183}~\textit{it maintains exploration freedom by providing a shortcut to the original problem}, allowing the model to implicitly correct or bypass unhelpful upstream reasoning. Formally:
\begin{equation}\label{eq:upstream_downstream}
\begin{aligned}
\mathrm{I}:\quad &s \sim \pi_{\boldsymbol{\theta}}(\cdot | q), \\
\mathrm{II}:\quad &\{o_i\}_{i=1}^{G} \sim \pi_{\boldsymbol{\theta}}(\cdot | [s, q]).
\end{aligned}
\end{equation}
where $q \in \mathcal{Q}$ denotes the problem, $s \in \mathcal{S}$ the reasoning segment, and $o \in \mathcal{O}$ the final output. We define reward signals accordingly: downstream reward $\mathcal{R}_{\mathrm{II}}(o, a) = 2 \cdot \mathbbm{1}[o = a] - 1$ measures final correctness, while upstream reward $\mathcal{R}_{\mathrm{I}}(s, a)$ implements the Monte Carlo estimation:
\begin{equation}
\label{eq:upstream_reward}
\mathcal{R}_{\mathrm{I}}(s, a) = \frac{1}{G} \sum_{i=1}^{G} \mathcal{R}_{\mathrm{II}}(o_i, a), \quad \text{where } o_i \sim \pi_{\boldsymbol{\theta}}(\cdot \mid [s, q]).
\end{equation}
This design achieves a favorable tradeoff: whereas GRPO requires $G$ complete responses per problem with no intermediate feedback, and full token-level Monte Carlo would require $K$ continuations for each token across all $G$ responses, our approach requires only $G$ samples total while providing one critical intermediate reward signal.

\textbf{Engineering Optimization.} A naive implementation would first generate $1$ upstream segment, then spawn $G$ downstream continuations, requiring two separate GPU batches. Although this saves the complexity of $(G-1)$ upstream segments compared to generating $G$ complete responses, the overhead of two separate batch dispatches actually increases overall training time significantly. Figure~\ref{fig:time_per_step} in \Cref{app:training_dynamics} illustrates the typical difference between two-batch approaches and our optimized single-pass implementation. Notably, Critique-GRPO, which first rolls out 7 direct responses then generates 1 self-critique response in a separate batch, exhibits substantially higher per-step training time due to this sequential dispatch overhead, with time complexity similar to our naive two-batch design. To achieve true computational parity with mainstream GRPO variants, we modify the vLLM~\cite{kwon2023vllm} inference engine to perform mid-generation KV cache rewriting within a single rollout pass. Concretely, all $G$ responses are generated in parallel from $q$, paused at the shared segmentation position $t_q$. At this point, each sequence $i$ maintains its own KV cache for segment $s_i$, while sharing the KV cache for prompt $q$ via pointer references. We then select the segment closest to the median deviation as the shared upstream prefix $s = s_{k^*}$, where $k^* = \arg\min_k |d(s_k) - \mathrm{median}(\{d(s_i)\})|$ and $d(s) = \frac{1}{|s|}\sum_{t} -\log \pi_{\boldsymbol{\theta}}(s_t | q, s_{<t})$ is the mean negative log-probability. The key operation is \textit{KV cache pointer redirection}: we redirect all $G$ sequences' segment KV cache pointers to the single selected $\text{KV}(s)$, then recompute the KV cache for the reordered prefix $[s, q]$. This enables all subsequent outcome tokens $o_i$ to be generated conditional on the identical prefix, ensuring valid Monte Carlo estimation while requiring only one shared copy of $\text{KV}(s)$ in GPU memory.

\subsection{Asymmetric Optimization Strategy}\label{sec:optimization}

Given a dataset $\mathcal{D}$ with ground-truth answers $a$, the skip-connected optimization objective maximizes the expected advantages at both stages:
\begin{equation}
\label{eq:objective}
\underset{\boldsymbol{\theta}}{\max} \;\mathbb{E}_{(q, a) \sim \mathcal{D}} \Big[ \mathcal{J}_{\mathrm{I}}(\boldsymbol{\theta}; q, a) + \mathcal{J}_{\mathrm{II}}(\boldsymbol{\theta}; q, s, a) \Big],
\end{equation}
where $\mathcal{J}_{\mathrm{I}}$ and $\mathcal{J}_{\mathrm{II}}$ are the upstream and downstream policy gradient objectives defined using advantages computed from rewards $\mathcal{R}_{\mathrm{I}}$ and $\mathcal{R}_{\mathrm{II}}$ respectively.

Given that the upstream response density is multiplicatively lower than that of the downstream, we employ an asymmetric optimization strategy using different algorithms for each phase, as detailed below.

\paragraph{Downstream Optimization: Group Relative Policy Optimization.}
For the downstream phase, we employ group-relative policy optimization with clipped importance sampling. For each augmented input $\tilde{q}_i = [s_i; q_i]$ (where $s_i$ denotes the selected segment for problem $q_i$, distinct from the $G-1$ discarded segments during parallel generation), we sample a group of $G$ responses and update the policy according to:
\begin{equation}
\label{eq:grpo}
\begin{aligned}
\mathcal{J}_{\mathrm{II}}(\boldsymbol{\theta}) &= \mathbb{E}_{(q, a) \sim \mathcal{D}, \{o_i\}_{i=1}^{G} \sim \pi_{\boldsymbol{\theta}_{\text{old}}}(\cdot|\tilde{q}_i)} \\
&\Bigg[ \frac{1}{\sum_i |o_i|} \sum_{i=1}^{G} \sum_{t=1}^{|o_i|} \texttt{\textbf{sg}}(\hat{r}_{i,t}) \, \hat{A}_{i} \log \pi_{\boldsymbol{\theta}}(o_{i,t} \mid \tilde{q}_i, o_{i,<t}) \Bigg],
\end{aligned}
\end{equation}
where $\texttt{\textbf{sg}}(\cdot)$ denotes the stop-gradient operation, $\hat{r}_{i,t} = \text{clip}(r_{i,t}, 1-\varepsilon, 1+\varepsilon)$ with $r_{i,t} = \frac{\pi_{\boldsymbol{\theta}}(o_{i,t} \mid \tilde{q}_i, o_{i,<t})}{\pi_{\boldsymbol{\theta}_{\text{old}}}(o_{i,t} \mid \tilde{q}_i, o_{i,<t})}$, and $\hat{A}_{i} = (R_i - \bar{R}) / \sigma_R$.

\paragraph{Upstream Optimization: Single-stream Policy Optimization.}
Since only one selected segment per problem participates in upstream optimization (with the remaining $G-1$ segments from parallel generation discarded after contributing to median-based selection), group-relative advantage estimation is infeasible. We adopt SPO~\cite{xu2025single}, which replaces the \textit{spatial} baseline, i.e., group mean from concurrent rollouts, with a \textit{temporal} baseline: a per-prompt tabular tracker $\hat{V}(q)$ updated via exponential moving average $\hat{V}(q) \leftarrow \hat{V}_{-1}(q) + \eta \cdot (\mathcal{R}_{\mathrm{I}}(s, a) - \hat{V}_{-1}(q))$, where $\hat{V}_{-1}(q)$ denotes the pre-update baseline (initialized to a global default, e.g., $0$, and immediately overwritten upon the first observation for each prompt), and the adaptive learning rate $\eta = (\rho N + 1)^{-1}$ depends on a KL-adaptive forgetting factor $\rho = 2^{-D_{\mathrm{KL}} / \tau_{\mathrm{half}}}$ and the discounted cumulative observation count $N$.\footnote{While the original SPO formulation assumes rewards in $[0, 1]$, our rewards $\mathcal{R}_{\mathrm{I}} \in [-1, +1]$ are affinely mapped to $[0, 1]$ via $(r+1)/2$ for tracker updates, then mapped back to $[-1, +1]$ when retrieving baselines. This preserves the EMA equivalence while maintaining numerical stability.} The advantage $A_i = \mathcal{R}_{\mathrm{I}}(s_i, a_i) - \hat{V}_{-1}(q_i)$ uses the pre-update baseline to ensure unbiasedness, then normalized across the batch: $\tilde{A}_i = (A_i - \mu_{\mathcal{B}}) / \sigma_{\mathcal{B}}$. The objective is:
\begin{equation}
\label{eq:spo}
\begin{aligned}
\mathcal{J}_{\mathrm{I}}(\boldsymbol{\theta}) &= \mathbb{E}_{(q, a) \sim \mathcal{D}, s_i \sim \pi_{\boldsymbol{\theta}_{\text{old}}}(\cdot | q)} \\
&\Bigg[ \frac{1}{|s_i|} \sum_{t=1}^{|s_i|} \min\big(r_{i,t}(\boldsymbol{\theta}) \, \tilde{A}_i,\, \hat{r}_{i,t} \, \tilde{A}_i\big) \Bigg],
\end{aligned}
\end{equation}
where $r_{i,t}(\boldsymbol{\theta}) = \pi_{\boldsymbol{\theta}}(s_{i,t} \mid q_i, s_{i,<t}) / \pi_{\boldsymbol{\theta}_{\text{old}}}(s_{i,t} \mid q_i, s_{i,<t})$ and $\hat{r}_{i,t} = \text{clip}(r_{i,t}, 1-\varepsilon, 1+\varepsilon)$.

\section{Experiments}
\label{sec:experiments}

\paragraph{Training Setup.}
We train \our{} on the dapo-math-17k dataset~\cite{yu2025dapo}, a curated mathematical reasoning dataset containing 17k problems spanning various difficulty levels. All models are trained for 500 optimization steps with a batch size of 128 prompts per step. We set the maximum prompt length to 1024 tokens and the maximum completion length to 3072 tokens for both training and evaluation. For fair comparison, all methods using group-relative advantage estimation adopt $G=8$ rollouts per prompt; with our engineering optimization, \our{} generates $G=8$ upstream segments in parallel, selects one via median-based selection, and continues with $G=8$ downstream responses within a single GPU batch. For SPO which does not use group-relative estimation, we scale the batch size by 8 to ensure equivalent computational cost; we also omit the warm-up phase in our SPO reproduction to avoid introducing a cost-misaligned initialization stage. We use Qwen2.5-Math-7B and Llama-3.2-3B-Instruct as base models. Complete hyperparameters and implementation details are provided in \Cref{app:experimental_setup}.

\subsection{Main Results}

We evaluate \our{} on diverse benchmarks covering In-Domain mathematical reasoning and Out-of-Domain general and code reasoning. For In-Domain evaluation, we conduct extensive experiments on mathematical benchmarks. For Out-of-Domain evaluation, we employ MMLU-Pro~\cite{wang2024mmlupro} to assess general reasoning capabilities and LiveCodeBench~\cite{jain2024livecodebench} to evaluate code generation skills. To reduce variance from stochastic sampling, we repeat evaluations 32 times for AIME24 and AIME25 due to their small dataset sizes, and 3 times for all other benchmarks. The main experimental results are summarized in \Cref{tab:performance}, where we compare \our{} against state-of-the-art baselines including GRPO~\cite{shao2024deepseekmath}, GSPO~\cite{zheng2025gspo}, SPO~\cite{xu2025single}, Critique-GRPO~\cite{zhang2025critique}, PRIME~\cite{cui2025prime}, DAPO~\cite{yu2025dapo}, SAPO~\cite{gao2025soft}, and CISPO~\cite{chen2025minimax}. These methods all rely solely on the policy model's own reasoning without external information, ensuring comparable inference costs, and represent a collection of recent influential approaches that have been widely adopted. Detailed experimental settings and hyperparameters are provided in \Cref{app:experimental_setup}.

\begin{table*}[!t]
\centering
\caption{Performance comparison on In-Domain (Math) and Out-of-Domain (General/Code) benchmarks. We evaluate on two base models: \textbf{Qwen2.5-Math-7B} and \textbf{Llama-3.2-3B-Instruct}. The best results for each model are \textbf{bolded}, and the second best are \underline{underlined}.}
\label{tab:performance}
\renewcommand\tabcolsep{2.5pt}
\renewcommand\arraystretch{1.0}
\resizebox{\textwidth}{!}{%
\begin{tabular}{l|cccccc|c@{\hspace{3pt}}c|c|ccccc|c@{\hspace{3pt}}c|c}
\Xhline{1.2pt}
\rowcolor{CadetBlue!20} 
    & \multicolumn{9}{c|}{\textbf{Qwen2.5-Math-7B}} & \multicolumn{8}{c}{\textbf{Llama-3.2-3B-Instruct}} \\
\rowcolor{CadetBlue!20} 
    & \multicolumn{6}{c|}{In-Domain (Math)} & \multicolumn{2}{c|}{Out-of-Domain} &  & \multicolumn{5}{c|}{In-Domain (Math)} & \multicolumn{2}{c|}{Out-of-Domain} & \\
\rowcolor{CadetBlue!20} 
\textbf{Method} & {\scriptsize\textbf{AIME24}} & {\scriptsize\textbf{AIME25}} & {\scriptsize\textbf{AMC}} & {\scriptsize\textbf{MATH}} & {\scriptsize\textbf{Minerva}} & {\scriptsize\textbf{Olympiad}} & {\scriptsize\textbf{MMLU}} & {\scriptsize\textbf{LiveCode}} & {\scriptsize\textbf{Avg.}} & {\scriptsize\textbf{AIME24}} & {\scriptsize\textbf{AIME25}} & {\scriptsize\textbf{AMC}} & {\scriptsize\textbf{MATH}} & {\scriptsize\textbf{Olympiad}} & {\scriptsize\textbf{MMLU}} & {\scriptsize\textbf{LiveCode}} & {\scriptsize\textbf{Avg.}} \\
\Xhline{1.2pt}
Base & $12.1$ & $5.4$ & $43.4$ & $58.1$ & $20.6$ & $34.1$ & $18.6$ & $13.8$ & $25.8$ & $3.4$ & $0.4$ & $20.1$ & $36.3$ & $18.1$ & $29.5$ & $12.4$ & $17.2$ \\
\rowcolor{gray!10}GRPO & $22.4$ & $10.1$ & $50.2$ & $63.8$ & $23.5$ & $35.8$ & $21.3$ & $14.6$ & $30.2$ & $4.7$ & $0.3$ & $24.8$ & $39.6$ & $19.8$ & $30.2$ & $13.1$ & $18.9$ \\
GSPO & $31.1$ & $14.8$ & $60.3$ & $74.2$ & $28.1$ & $43.6$ & $25.2$ & $16.2$ & $36.7$ & $6.9$ & $0.5$ & $30.4$ & $\underline{42.1}$ & $22.3$ & $\textbf{30.5}$ & $\underline{13.8}$ & $21.0$ \\
\rowcolor{gray!10}SPO & $28.8$ & $13.6$ & $56.8$ & $65.9$ & $29.2$ & $40.1$ & $24.1$ & $15.7$ & $34.3$ & $5.8$ & $0.4$ & $26.1$ & $39.2$ & $20.3$ & $29.8$ & $12.9$ & $19.2$ \\
Critique$^\dagger$ & $28.2$ & $13.2$ & $54.2$ & $63.2$ & $22.1$ & $38.7$ & $22.8$ & $15.2$ & $32.2$ & N/A & N/A & N/A & N/A & N/A & N/A & N/A & N/A \\
\rowcolor{gray!10}SAPO & $32.0$ & $14.2$ & $64.7$ & $76.1$ & $\underline{33.8}$ & $46.5$ & $\textbf{28.7}$ & $16.5$ & $39.1$ & $7.9$ & $\textbf{3.3}$ & $\textbf{38.1}$ & $42.0$ & $21.2$ & $\underline{30.4}$ & $12.6$ & $22.2$ \\
PRIME & $30.1$ & $15.7$ & $63.9$ & $71.3$ & $30.8$ & $43.9$ & $25.0$ & $\underline{18.9}$ & $37.5$ & $5.4$ & $0.8$ & $29.7$ & $34.7$ & $20.8$ & $30.1$ & $13.2$ & $19.2$ \\
\rowcolor{gray!10}DAPO & $33.5$ & $15.9$ & $\underline{69.2}$ & $75.8$ & $30.4$ & $47.2$ & $27.1$ & $17.0$ & $39.5$ & $\textbf{13.8}$ & $0.4$ & $36.5$ & $41.1$ & $24.3$ & $30.2$ & $12.7$ & $\underline{22.7}$ \\
CISPO & $\underline{34.8}$ & $\underline{17.8}$ & $68.1$ & $\underline{79.4}$ & $32.6$ & $\underline{49.3}$ & $27.8$ & $17.4$ & $\underline{40.9}$ & $10.8$ & $0.4$ & $25.3$ & $35.8$ & $\underline{24.4}$ & $29.9$ & $12.1$ & $19.8$ \\
\rowcolor{yellow!10}\textbf{\our{} (Ours)} & $\textbf{35.7}$ & $\textbf{18.7}$ & $\textbf{71.4}$ & $\textbf{80.8}$ & $\textbf{34.7}$ & $\textbf{51.0}$ & $\underline{28.4}$ & $\textbf{19.6}$ & $\textbf{42.5}$ & $\underline{14.7}$ & $\underline{1.1}$ & $\underline{37.9}$ & $\textbf{44.8}$ & $\textbf{26.3}$ & $30.1$ & $\textbf{14.1}$ & $\textbf{24.1}$ \\
\Xhline{1.2pt}
\end{tabular}%
}
{\scriptsize $^\dagger$The original Critique-GRPO uses external strong models (e.g., GPT-4o) to generate critiques. To align computational costs across all methods, we implement a self-critique variant where the policy model generates critiques itself, rather than the external-critique approach advocated in the original paper. We use 7 direct rollouts plus 1 self-reflection rollout (8 total, matching other methods). We failed to obtain stable training curves on Llama-3.2-3B-Instruct.}
\end{table*}

As shown in \Cref{tab:performance}, \our{} achieves strong improvements on Qwen2.5-Math-7B, where it leads in nearly all benchmarks with an average score of 42.5\%, surpassing the second-best method CISPO (40.9\%) by 1.6 percentage points and outperforming vanilla GRPO (30.2\%) by 12.3 percentage points. On Llama-3.2-3B-Instruct, \our{} yields the best overall average of 24.1\%, exceeding the second-best DAPO (22.7\%) by 1.4 percentage points and GRPO (18.9\%) by 5.2 percentage points, despite competitive per-task variation. On In-Domain mathematical benchmarks, \our{} substantially outperforms strong baselines including CISPO and DAPO, demonstrating that the skip-connected architecture effectively integrates fine-grained rewards with group-relative estimation. More importantly, these gains transfer to Out-of-Domain tasks: on MMLU-Pro and LiveCodeBench, \our{} maintains its advantage, demonstrating that the hierarchical reasoning structure generalizes beyond pure mathematics to broader reasoning domains including code generation.

\begin{figure}[t]
    \centering
    \includegraphics[width=\linewidth]{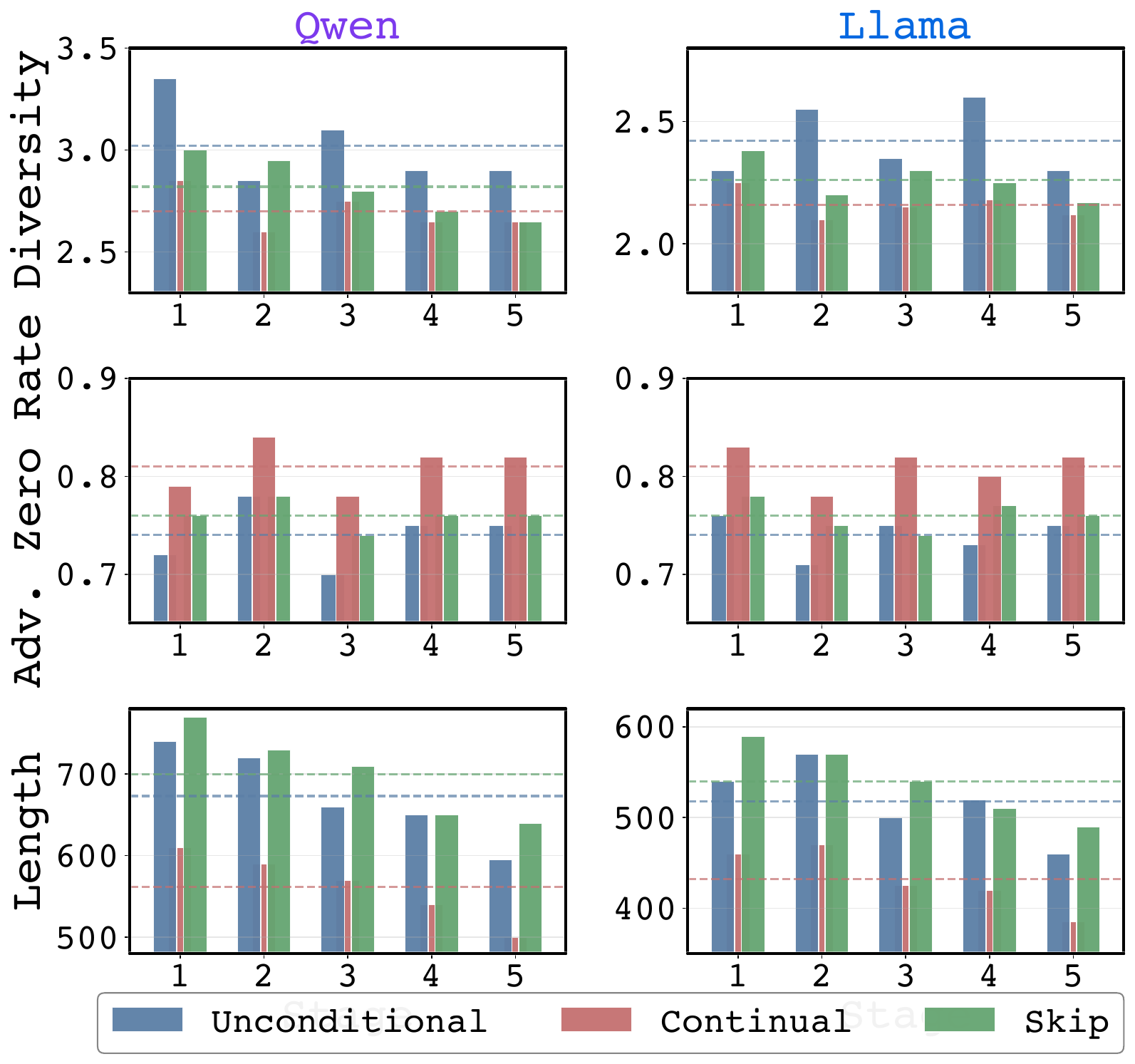}
    \caption{Upstream impact analysis comparing three conditioning strategies across Qwen and Llama models. \textit{Top}: Diversity (average distinct answers per 8 rollouts); \textit{Middle}: Advantage zero rate (probability of homogeneous outcomes); \textit{Bottom}: Response length (excluding prefix). Continual conditioning leads to diversity collapse and elevated advantage zero rate, while Skip maintains properties comparable to Unconditional.}
    \label{fig:upstream_impact}
\end{figure}

\paragraph{Shortcut Validation.}
\label{sec:design_validation}
The skip connection $[s, q]$ concatenates the upstream segment and the original problem to balance upstream influence and downstream exploration freedom. We compare three conditioning strategies: \textbf{Unconditional} generating from $q$ alone, \textbf{Continual} using $s$ as a hard prefix, and \textbf{Skip} (ours) conditioning on $[s, q]$. Using base models (Qwen2.5-Math-7B and Llama-3.2-3B-Instruct) without any RL training, we first generate complete responses for each problem, then extract segments $s$ at six relative positions spanning from $\frac{1}{6}$ to $\frac{1}{2}$ of the total response length. We then sample $G=8$ downstream continuations under each conditioning strategy on a mixture of all mathematical evaluation benchmarks (AIME24, AIME25, AMC, MATH, Minerva, and Olympiad). We measure three metrics: \textbf{Diversity} as the average number of distinct final answers among 8 rollouts per problem; \textbf{Advantage Zero Rate} as the probability that all 8 rollouts yield identical outcomes (all correct or all incorrect), which collapses the advantage signal to zero; and \textbf{Response Length} as the generation length excluding the conditioning prefix. As shown in \Cref{fig:upstream_impact}, Continual conditioning progressively constrains exploration as segments extend, resulting in lower diversity, higher advantage zero rate, and shorter generation length. In contrast, Skip maintains diversity and advantage zero rate comparable to Unconditional, demonstrating that the skip-connected structure effectively preserves downstream exploration freedom while incorporating upstream guidance.

\subsection{Ablation Studies}
\label{sec:ablation}

We conduct ablation studies to validate the effectiveness of key components in \our{}. We compare \our{} against four variants: \ding{182}~\textbf{w/o Skip}: The upstream segment $s$ is directly used as a prefix for downstream generation without the skip connection to the original problem. \ding{183}~\textbf{w/o MC Estimation}: The upstream stage receives rewards from its own complete generation without aggregating downstream outcomes. \ding{184}~\textbf{Split@Fixed}: Fixed split position alternating between $\frac{1}{6}$ and $\frac{1}{2}$ across training steps, replacing our adaptive sampling from $[\frac{1}{6}L_q, \frac{1}{2}L_q]$. \ding{185}~\textbf{Selection:Extreme}: Select the upstream $s$ with minimum or maximum negative log-probability alternating across steps, instead of the median. The results are summarized in \Cref{tab:ablation}.

\begin{table}[t]
\centering
\caption{Ablation study on In-Domain benchmarks comparing different structural settings. We report accuracy (\%) for Qwen2.5-Math-7B. The best results are \textbf{bolded}.}
\label{tab:ablation}
\renewcommand\tabcolsep{4.5pt}
\renewcommand\arraystretch{1.05}
\resizebox{0.95\linewidth}{!}{%
\begin{tabular}{l|cccccc}
\Xhline{1.2pt}
\rowcolor{CadetBlue!20} 
\textbf{Setting} & \textbf{AIME24} & \textbf{AIME25} & \textbf{AMC} & \textbf{MATH} & \textbf{Minerva} & \textbf{Olympiad} \\
\Xhline{1.2pt}
w/o Skip & $32.8$ & $15.4$ & $67.8$ & $74.6$ & $29.8$ & $46.3$ \\
w/o MC Estimation & $33.6$ & $17.4$ & $68.6$ & $78.3$ & $32.9$ & $48.7$ \\
Split@Fixed & $29.5$ & $14.1$ & $58.2$ & $67.3$ & $30.0$ & $41.5$ \\
Selection:Extreme & $28.2$ & $13.3$ & $55.9$ & $65.1$ & $28.6$ & $39.4$ \\
\rowcolor{gray!10}\textbf{Ours} & \textbf{$35.7$} & \textbf{$18.7$} & \textbf{$71.4$} & \textbf{$80.8$} & \textbf{$34.7$} & \textbf{$51.0$} \\
\Xhline{1.2pt}
\end{tabular}%
}
\end{table}

The results in \Cref{tab:ablation} validate all key components. \textbf{w/o Skip} leads to a significant performance drop: without the shortcut connection to the original problem, downstream responses sampled along the same upstream path explore a constrained space, substantially increasing the probability of homogeneous outcomes and advantage collapse to zero. \textbf{w/o MC Estimation} also underperforms: when the upstream stage receives rewards only from its own complete generation rather than aggregating downstream outcomes, it cannot learn reasoning patterns that specifically facilitate downstream solving. For segmentation position, \textbf{Split@Fixed} shows degraded performance: alternating between fixed positions loses the benefit of adaptive sampling, as splitting too early provides insufficient upstream context while splitting too late reduces downstream exploration freedom. Our adaptive sampling from $[\frac{1}{6}L_q, \frac{1}{2}L_q]$ dynamically balances both considerations. For upstream selection, \textbf{Selection:Extreme} alternating between min and max perplexity underperforms because extreme selections either represent overconfident but potentially flawed reasoning (min) or uncertain responses lacking useful guidance (max). The median strategy avoids both extremes, selecting a representative upstream segment that facilitates diverse downstream exploration.

\subsection{Implicit Advantage Analysis}
\label{sec:advanced_analysis}

\begin{figure}[t]
    \centering
    \includegraphics[width=\linewidth]{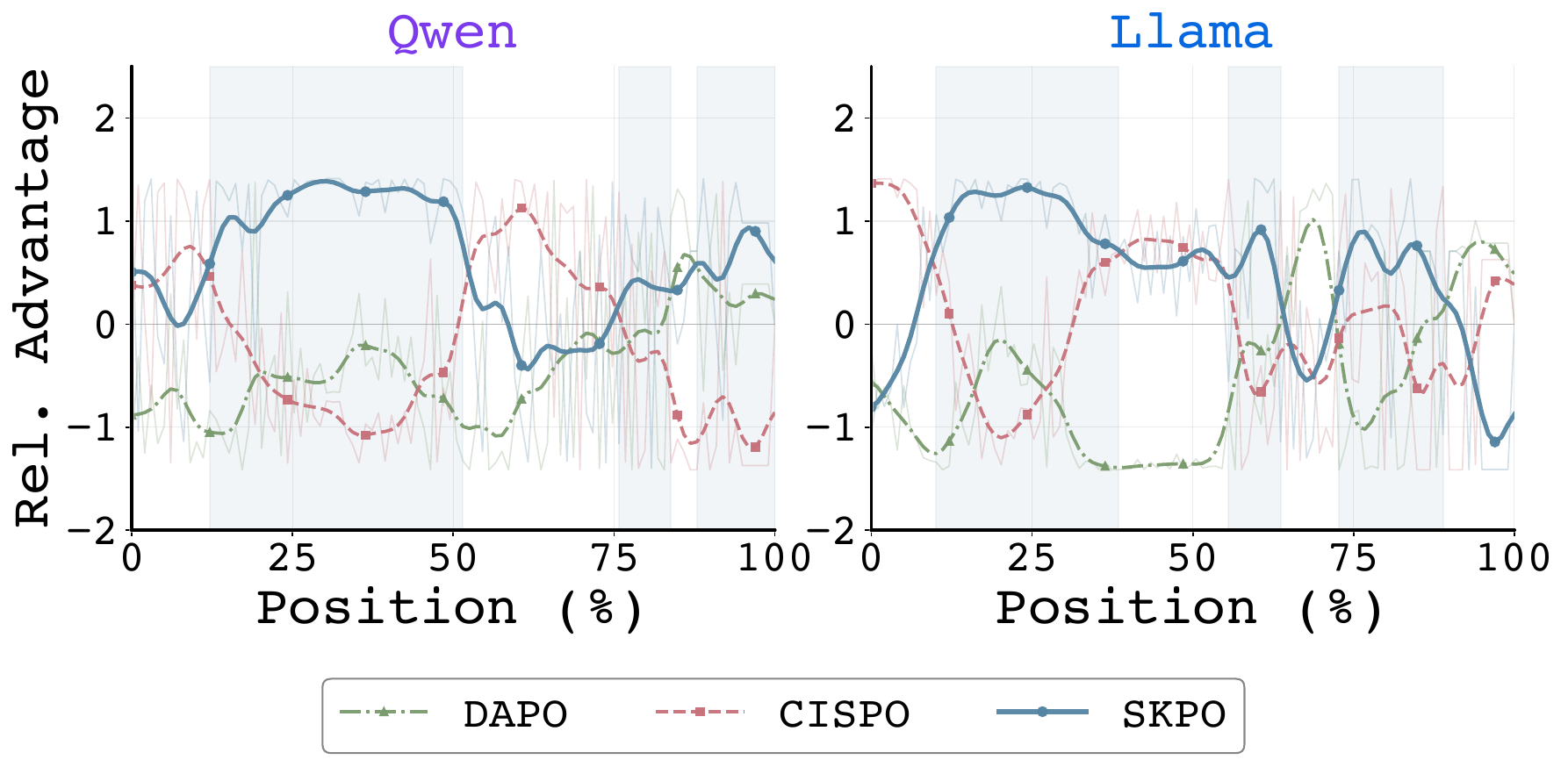}
    \caption{Segment-level implicit advantage analysis using GPT-5-nano as external evaluator across both Qwen and Llama models. For each method's correct responses, we split into 100 segments by relative position and estimate rewards via Monte Carlo continuation sampling. Inter-method relative advantages are computed at each position. \our{} consistently maintains higher advantages in the 20--50\% position range, indicating superior intermediate reasoning quality during the problem-setup phase.}
    \label{fig:token_reward_analysis}
\end{figure}

Following the Monte Carlo trajectory sampling approach for implicit reward estimation, we design an analysis to reveal how different optimization objectives shape intermediate reasoning quality. We collect all correct responses from \our{}, CISPO, and DAPO across all mathematical test benchmarks (AIME24, AIME25, AMC, MATH, Minerva, and Olympiad), splitting each into 100 segments based on relative token position. We select CISPO and DAPO as comparison baselines because they achieve relatively strong performance among all methods in our main experiments. For each segment, we use GPT-5-nano as the external evaluator to perform Monte Carlo continuation sampling 8 times (see \Cref{app:prompt} for the prompt template), obtaining segment-level reward estimates that reflect ``how easy is it to reach a correct answer from this intermediate state.'' We then compute \textit{inter-method relative advantages} at each position following $\hat{A}_{\text{method}} = (R_{\text{method}} - \bar{R}) / \sigma_R$, where the comparison group consists of all methods at the same relative position for the same problem. We choose GPT-5-nano because its cost is acceptable and its mathematical reasoning capability is sufficient for producing meaningful visualizations.

\paragraph{Why External Evaluator?} A natural alternative would be mutual evaluation among trained models. However, since rewards derive from Monte Carlo sampling, models trained on identical data with only algorithmic differences exhibit similar continuation distributions, making inter-method differences difficult to detect without prohibitively large sample sizes. Using a substantially stronger external model addresses this: its higher capability provides a wider dynamic range in reward estimates, amplifying subtle quality differences in intermediate reasoning that would otherwise be obscured by sampling variance.

As visualized in \Cref{fig:token_reward_analysis}, responses from \our{} consistently maintain significantly higher advantages in the early-to-middle stages (20--50\% position range) compared to those from CISPO and DAPO, across both model families. This cross-model consistency reinforces the finding: \our{} establishes superior structural foundations during the critical problem-setup phase, where our hierarchical optimization explicitly provides learning signals. The convergence of all curves toward zero at trajectory end is expected, as outcome correctness becomes determined and intermediate advantages diminish.

\section{Conclusion}
\label{sec:conclusion}

We identify a key challenge in integrating fine-grained rewards with group-relative optimization: insufficient Monte Carlo sampling for token-level reward estimation degrades performance below outcome-only baselines. To address this, we propose \our{}, which partitions reasoning into upstream and downstream phases via a skip-connected architecture. The upstream phase generates early-stopped reasoning segments; the downstream phase performs trajectory sampling from the skip connection of the segment and original problem, preserving upstream influence while maintaining exploration freedom. We employ asymmetric optimization: single-stream algorithms for the sample-constrained upstream phase and group-relative algorithms for the downstream phase. Experiments show \our{} outperforms strong baselines on In-Domain mathematics and generalizes to Out-of-Domain tasks. Our off-policy continuation analysis reveals an \keyword{implicit} \keyword{advantage} phenomenon: \our{}-generated segments exhibit systematically higher advantages in intermediate steps when comparing correct responses across methods, indicating more robust forward reasoning patterns.

\paragraph{Limitations.}
Our study still has several limitations. First, the main training experiments are conducted on 3B and 7B models with mathematical RLVR data, so broader validation on larger scales and non-math training domains remains future work. Second, \our{} currently instantiates Monte Carlo sampling at a single split point sampled from a restricted range; richer multi-split or later-split designs may further improve granularity at higher cost. Third, our implicit-advantage analysis relies on an external evaluator with instruction-based continuation rather than raw prefix completion, so the resulting reward estimates should be interpreted comparatively rather than as absolute values. Finally, our single-pass KV-cache rewriting is an efficiency optimization rather than a correctness requirement, but reproducing the most efficient version still requires custom inference-engine support.

Future work can explore later split positions with higher resampling costs for finer granularity, and integration with multi-stage RL for agentic tasks.

\section*{Impact Statement}

This paper presents work whose goal is to advance the field of 
Deep Learning, specifically in reinforcement learning for 
large language models. The proposed method aims to improve sample 
efficiency and training stability, which could reduce computational 
costs and environmental impact. However, there are potential risks 
if more capable reasoning models are used for malicious purposes. 
We encourage responsible development and deployment of such technologies.


\bibliography{references}
\bibliographystyle{icml2026}

\newpage
\appendix
\onecolumn
\section{Why Fine-Grained Rewards Fail with Group-Relative Estimation}
\label{app:mc_variance}

\begin{figure}[htbp]
    \centering
    \includegraphics[width=0.9\linewidth]{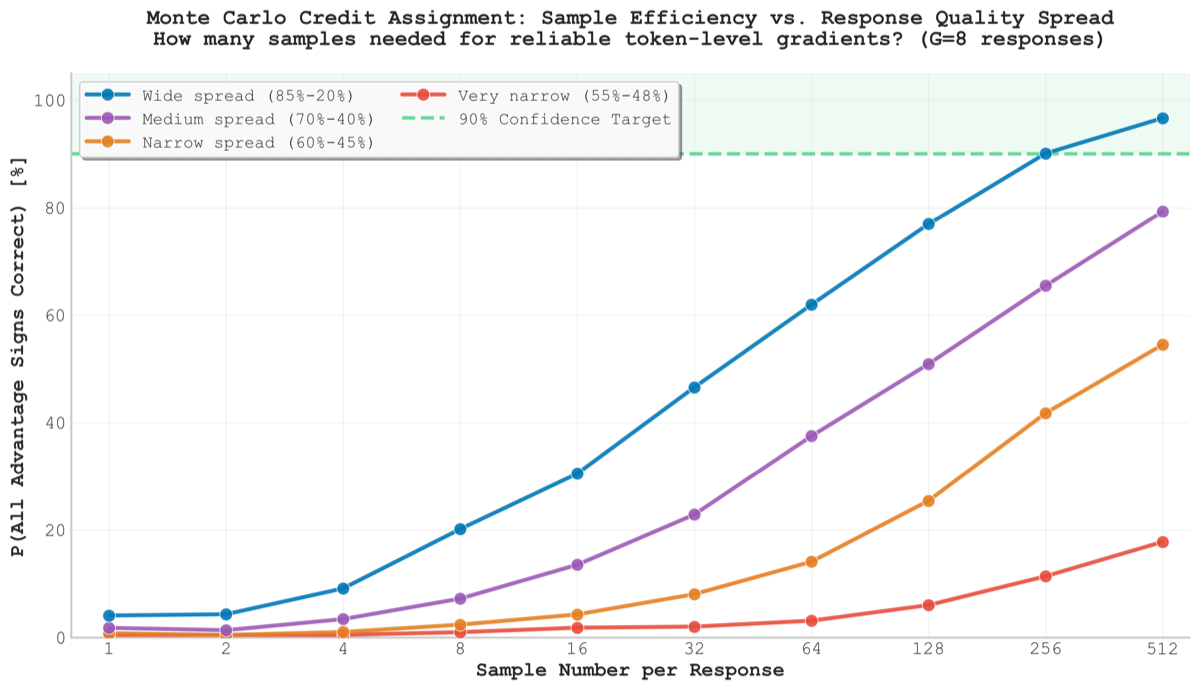}
    \caption{Probability of correct advantage signs for all $G=8$ samples across varying group spreads. Reliable credit assignment for narrow spreads requires prohibitive sample counts ($N>512$).}
    \label{fig:mc_sample_vs_correctness}
\end{figure}

As claimed in \Cref{sec:introduction}, directly integrating fine-grained rewards with group-relative advantage estimation leads to a fundamental sign inconsistency problem: with limited Monte Carlo samples, estimated advantages frequently have incorrect signs, causing the optimization to reinforce wrong behaviors or penalize correct ones. This appendix provides simulation evidence quantifying this phenomenon. The severity of this problem explains why existing methods \textsl{\texttt{circumvent}} rather than solve it. For example, PRIME~\citep{cui2025prime} aggregates token rewards to response-level before computing group-relative advantages, while VSRM~\citep{yue2025vsrm} abandons group-relative estimation entirely in favor of PPO. \our{} takes a different approach: \textsl{\texttt{structurally partitioning}} the problem so that upstream optimization avoids group-relative computation while downstream optimization has sufficient samples for stable estimation.

\subsection{Position-Dependent Variance}

The effectiveness of group-relative advantage estimation varies dramatically with trajectory position. For very early tokens (problem restatement and initial setup), the reward is largely determined by the model's own policy distribution rather than the specific tokens generated, as the reasoning space has not yet been meaningfully pruned. Trajectory sampling from these positions yields high variance, but this variance is \textsl{\texttt{symmetric}} across methods, explaining why our implicit advantage analysis shows similar rewards at the 0--20\% position range. Intermediate tokens (the decisive reasoning steps) are where methods diverge. Quality differences emerge as some reasoning paths make productive progress while others take detours or make errors. However, Monte Carlo estimation at these positions suffers from prohibitive variance: distinguishing a 55\% success probability prefix from a 60\% one requires hundreds of samples. This is precisely where \our{}'s implicit advantage emerges, as our upstream-downstream structure provides stable optimization signals for these intermediate segments. For terminal tokens (final computation and answer extraction), the outcome is semantically determined; sampling variance is minimal and single samples provide reliable gradients, so all methods perform comparably.

\begin{figure}[htbp]
    \centering
    \begin{subfigure}{0.9\linewidth}
        \includegraphics[width=\linewidth]{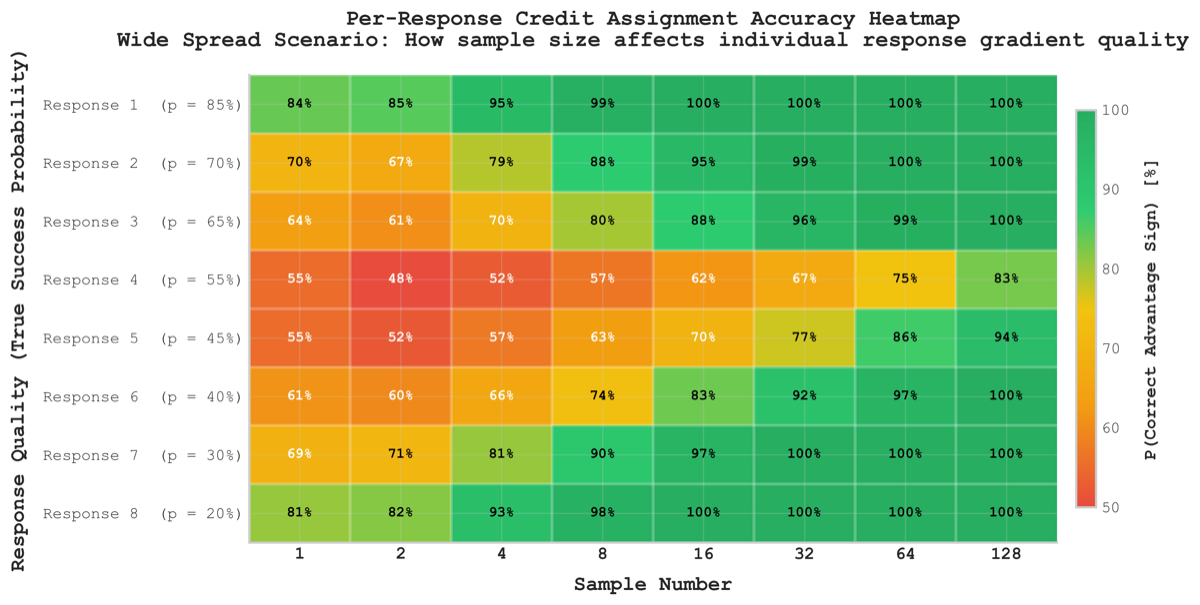}
        \caption{Credit assignment accuracy heatmap. Prefixes near the group mean are most susceptible to sign errors.}
        \label{fig:mc_per_response}
    \end{subfigure}
    
    \vspace{10pt}
    
    \begin{subfigure}{0.9\linewidth}
        \includegraphics[width=\linewidth]{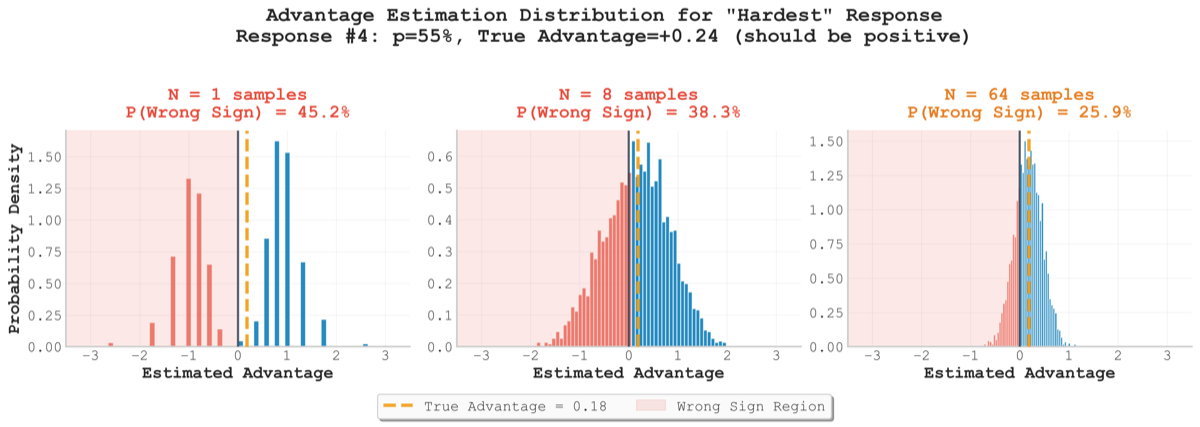}
        \caption{Distribution of advantage estimates for a prefix with $p=0.55$. Even with $N=8$, substantial overlap between positive and negative estimates remains.}
        \label{fig:mc_advantage_dist}
    \end{subfigure}
    \caption{Monte Carlo variance analysis. (a) Classification accuracy for individual samples. (b) Distribution shift and sign volatility under sampling noise.}
    \label{fig:mc_detail_combined}
\end{figure}

\subsection{Simulation Setup and Results}

We simulate a group of $G=8$ trajectories where each prefix has a true success probability $p_i$. Advantages are estimated by sampling $N$ continuations per prefix and applying group-relative normalization. An estimate is considered correct if the signs of all $G$ advantages match their ground-truth signs. Simulation results in \Cref{fig:mc_sample_vs_correctness,fig:mc_detail_combined} demonstrate that sample efficiency degrades rapidly as group quality spread narrows. When reasoning steps provide only incremental improvements, even $N=512$ samples fail to yield 50\% probability of correct sign assignment for the entire group. Individual prefix analysis in \Cref{fig:mc_per_response} confirms that prefixes near the group mean are most vulnerable to noise-induced sign flips, which are precisely the prefixes that represent incremental but genuine reasoning progress.

\subsection{Correctness Distribution Analysis}
\label{app:correctness_distribution}

\begin{figure}[htbp]
    \centering
    \begin{subfigure}{0.48\linewidth}
        \includegraphics[width=\linewidth]{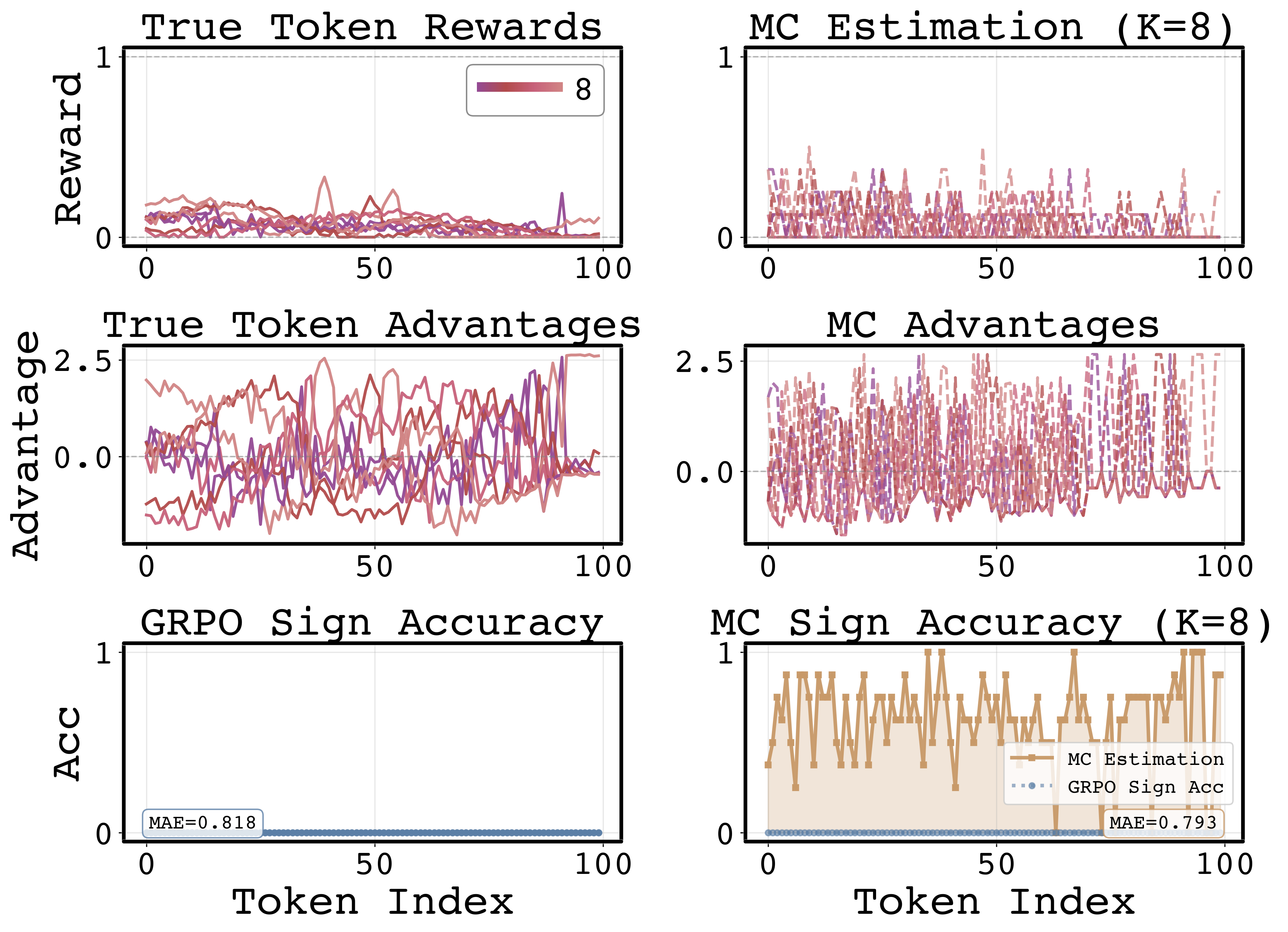}
        \caption{All incorrect (0/8)}
        \label{fig:correctness_0_8}
    \end{subfigure}
    \hfill
    \begin{subfigure}{0.48\linewidth}
        \includegraphics[width=\linewidth]{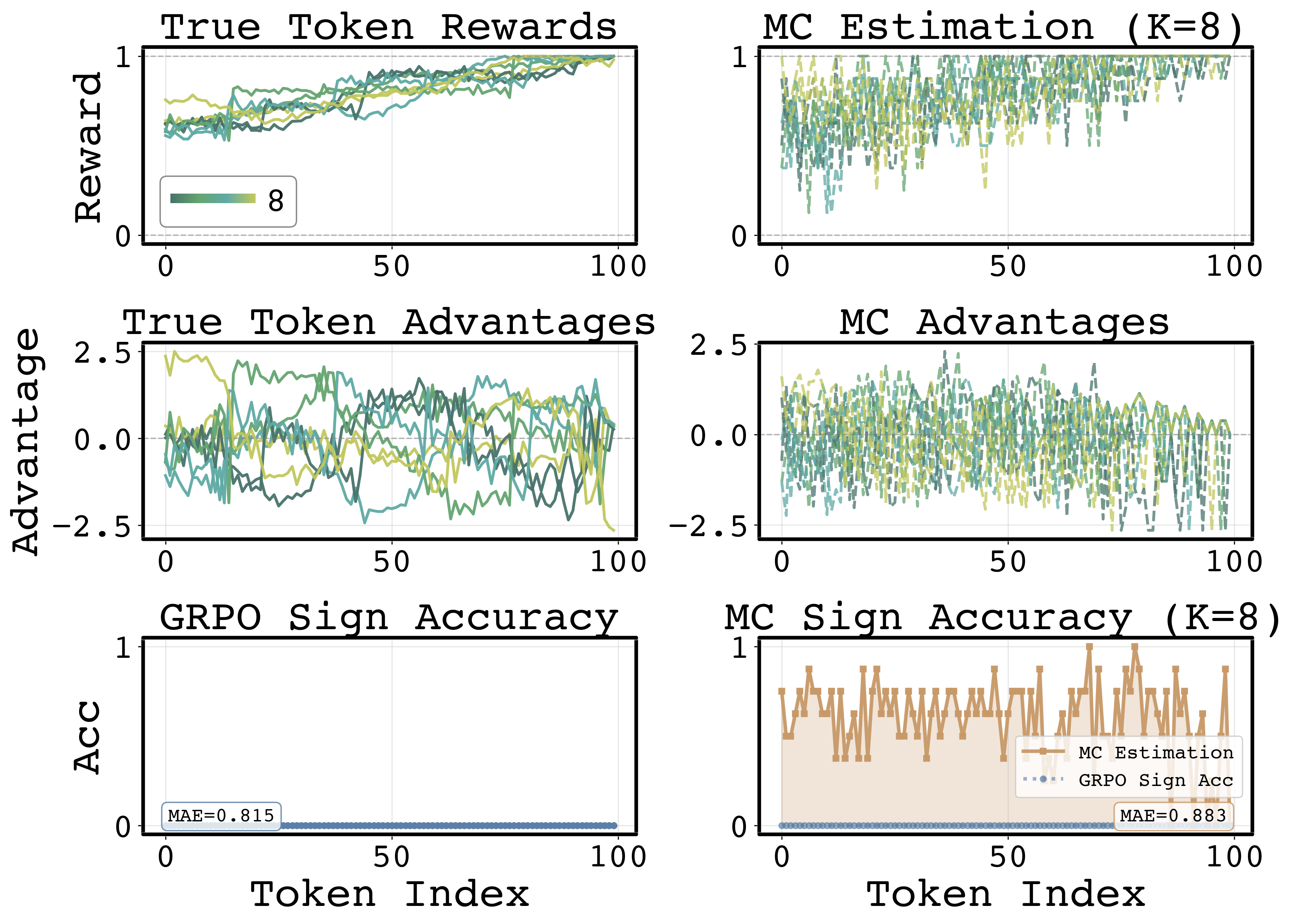}
        \caption{All correct (8/8)}
        \label{fig:correctness_8_8}
    \end{subfigure}
    
    \vspace{5pt}
    
    \begin{subfigure}{0.48\linewidth}
        \includegraphics[width=\linewidth]{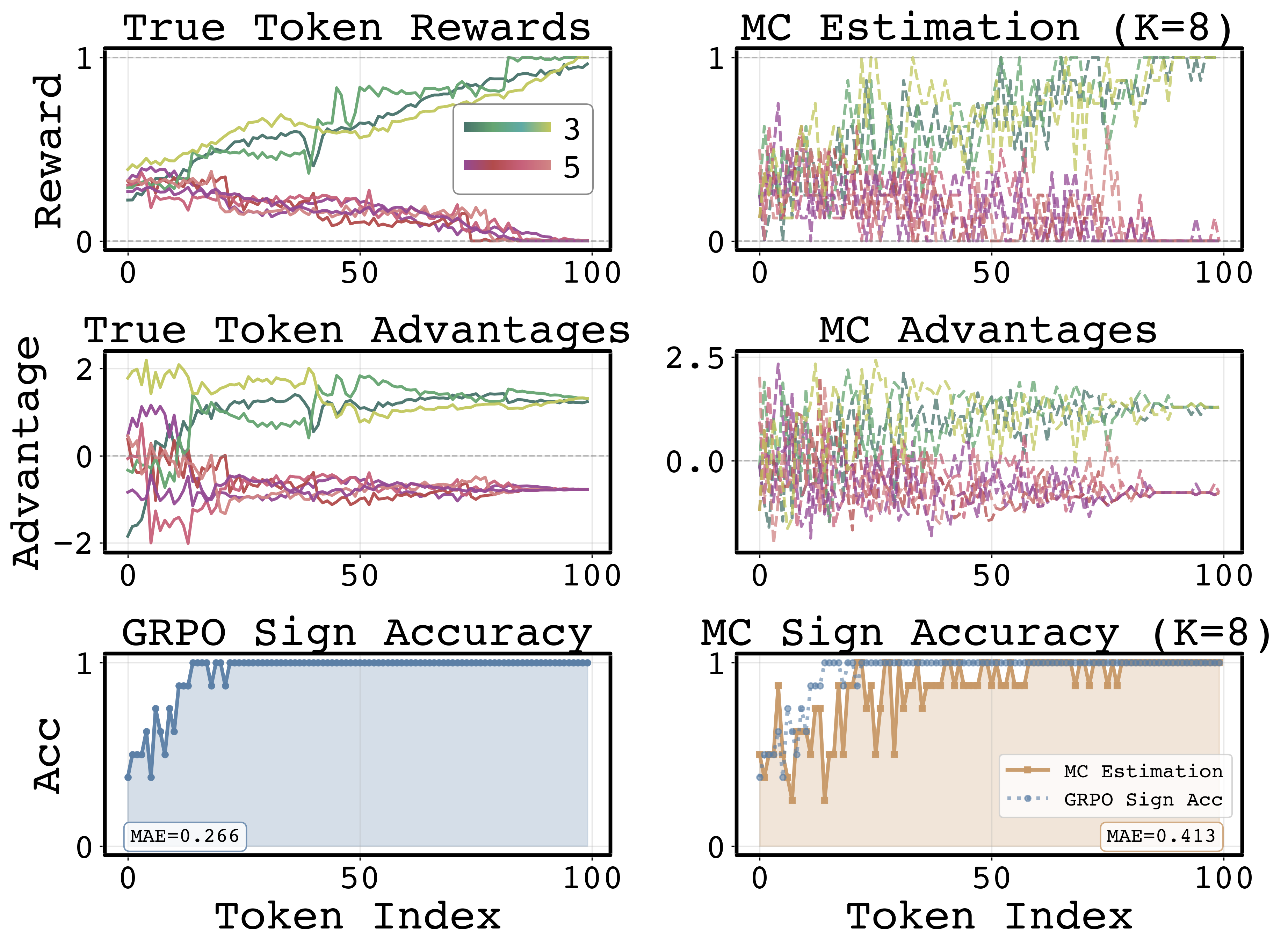}
        \caption{Mixed (3/8 correct)}
        \label{fig:correctness_3_8}
    \end{subfigure}
    \hfill
    \begin{subfigure}{0.48\linewidth}
        \includegraphics[width=\linewidth]{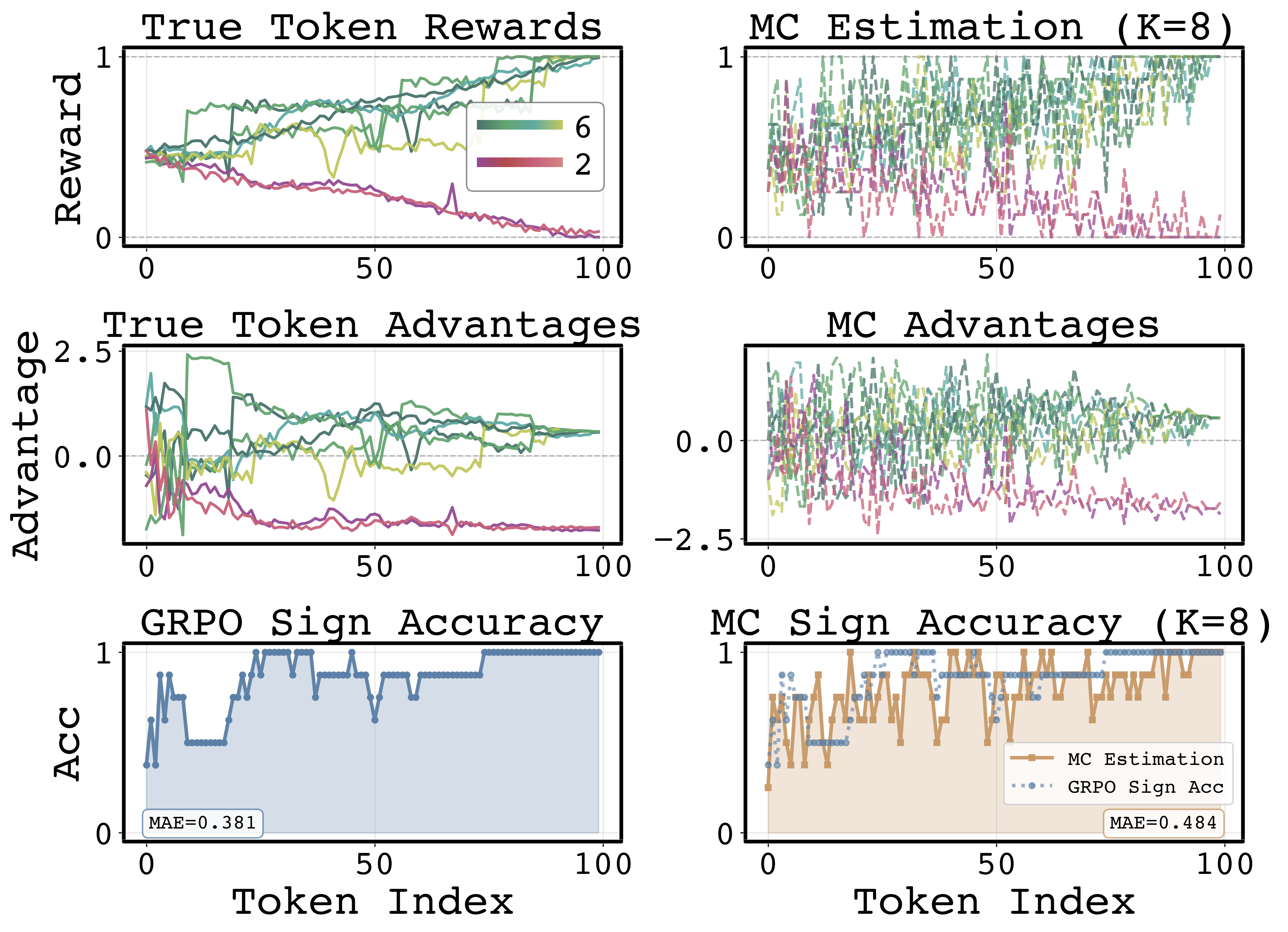}
        \caption{Mixed (6/8 correct)}
        \label{fig:correctness_6_8}
    \end{subfigure}
    \caption{Advantage estimation accuracy across different correctness distributions. \textbf{Key observation}: Only extreme cases (all correct or all incorrect) achieve reliable sign consistency. Mixed distributions suffer from severe sign errors throughout the trajectory, with accuracy often below 50\%. Additional examples (1/8, 2/8, 4/8, 6/8, 7/8) are available in \texttt{figures/mc/}.}
    \label{fig:correctness_dist_examples}
\end{figure}

To further illustrate the sign inconsistency problem, we systematically examine how advantage estimation accuracy varies across different correctness distributions within a group of $G=8$ responses. We sample random trajectories and categorize them by the number of correct responses (from 0/8 to 8/8), visualizing the true token rewards, advantages, and estimation accuracy for each distribution type. Representative examples are shown in \Cref{fig:correctness_dist_examples}.

The results reveal a striking pattern: \textbf{reliable advantage estimation occurs only in extreme cases}. When all responses converge to the same outcome (0/8 or 8/8), both GRPO and MC estimation achieve near-perfect sign accuracy, as there is no meaningful quality distinction to be made within the group. However, this apparent success is misleading: although sign accuracy appears high, the MAE between MC and true advantages remains comparable to GRPO in these degenerate cases, since all advantages are effectively zero and provide no learning signal.

In contrast, the informative mixed cases (1/8 through 7/8) exhibit persistent sign inconsistency problems:
\begin{itemize}[leftmargin=*]
    \item \textbf{Early tokens}: Sign accuracy frequently drops below 50\%, meaning random guessing would perform equally well.
    \item \textbf{Trajectory crossings}: When reward trajectories of different-outcome responses cross (common in early-to-mid reasoning), MC estimation cannot reliably distinguish which response is on a better path.
    \item \textbf{Narrow spreads}: Cases near 4/8 (balanced correct/incorrect) show the worst estimation quality, as the group quality spread is minimal.
\end{itemize}

\begin{figure}[htbp]
    \centering
    \includegraphics[width=\linewidth]{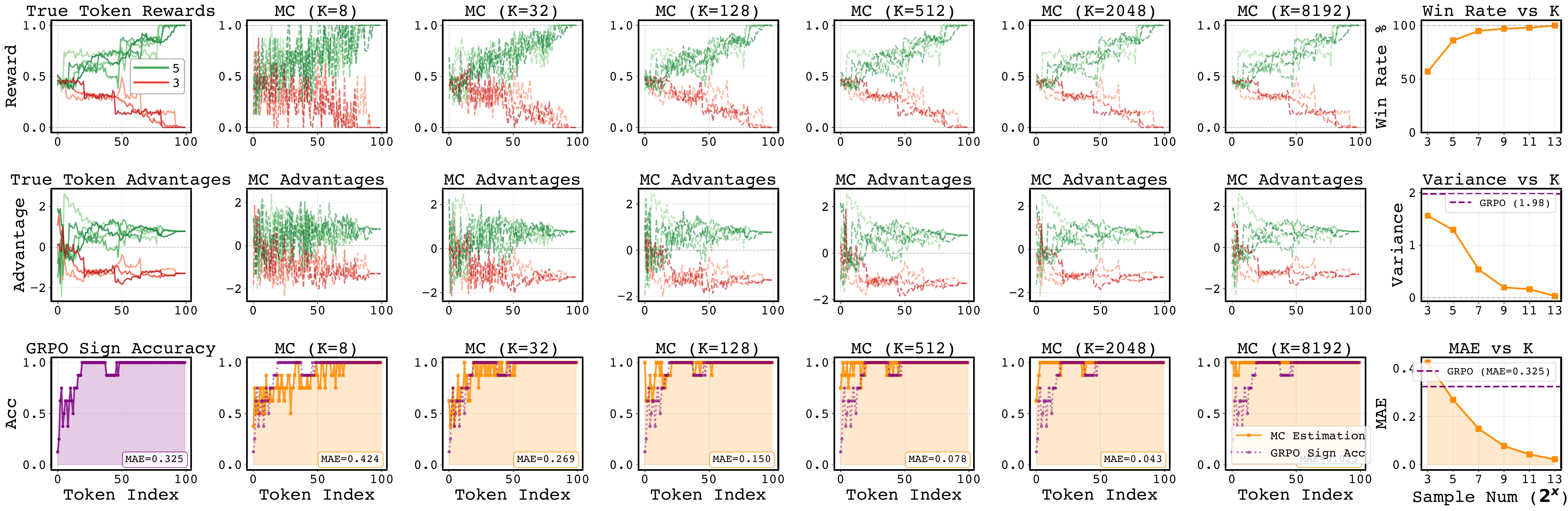}
    \caption{MC estimation accuracy across different sampling budgets ($K=8$ to $K=8192$). Left column shows true token rewards and advantages. Middle columns show MC estimates at each $K$ value. Right column summarizes win rate, variance, and MAE trends. With sufficient samples ($K \geq 512$), MC consistently outperforms GRPO, but the required sampling cost (up to 1000$\times$) is impractical for real-world training.}
    \label{fig:mc_multi_k}
\end{figure}

However, as shown in \Cref{fig:mc_multi_k}, with sufficient sampling budget, MC estimation \textit{does} eventually surpass GRPO. The figure demonstrates that as $K$ increases from 8 to 8192, MC sign accuracy improves steadily, and the Mean Absolute Error (MAE) between estimated and true advantages decreases substantially. At $K=8192$, MC achieves near-perfect sign accuracy with MAE approaching zero, significantly outperforming GRPO's constant-advantage approximation. This confirms that the theoretical advantage of fine-grained credit assignment is real; the challenge lies purely in the prohibitive sampling cost required to realize it.

This analysis underscores our central claim: fine-grained Monte Carlo estimation with practical sampling budgets only works reliably when it provides no useful signal, and fails precisely when fine-grained credit assignment would be most valuable.

\subsection{How \our{} Structurally Partitions the Problem}

These findings explain why \our{}'s architectural design is effective. In the upstream stage, by using single-stream policy optimization (SPO) with temporal baselines instead of group-relative advantages, \our{} entirely avoids the sign inconsistency problem for intermediate reasoning segments. The temporal baseline $\hat{V}(q)$ provides stable value estimates without requiring multiple concurrent samples. In the downstream stage, with $G$ rollouts per upstream segment, group-relative estimation becomes reliable. The downstream responses are closer to terminal tokens where outcome variance is naturally lower, and the group size provides sufficient samples for stable advantage computation. This structural partitioning (not circumvention) enables \our{} to genuinely optimize at the segment level, which is why our advanced off-policy analysis can reveal implicit advantage in intermediate reasoning steps that other methods cannot observe.

\section{Training Dynamics Analysis}
\label{app:training_dynamics}

We analyze the optimization behavior across different training configurations, examining convergence patterns, policy stability indicators, and computational efficiency metrics throughout the training process. All training curves presented in this section are from experiments on \textbf{Qwen2.5-Math-7B} base model. For visual clarity, we select a representative subset of methods in each figure rather than displaying all baselines.

\subsection{Performance Metrics}

\begin{figure}[htbp]
    \centering
    \includegraphics[width=0.85\linewidth]{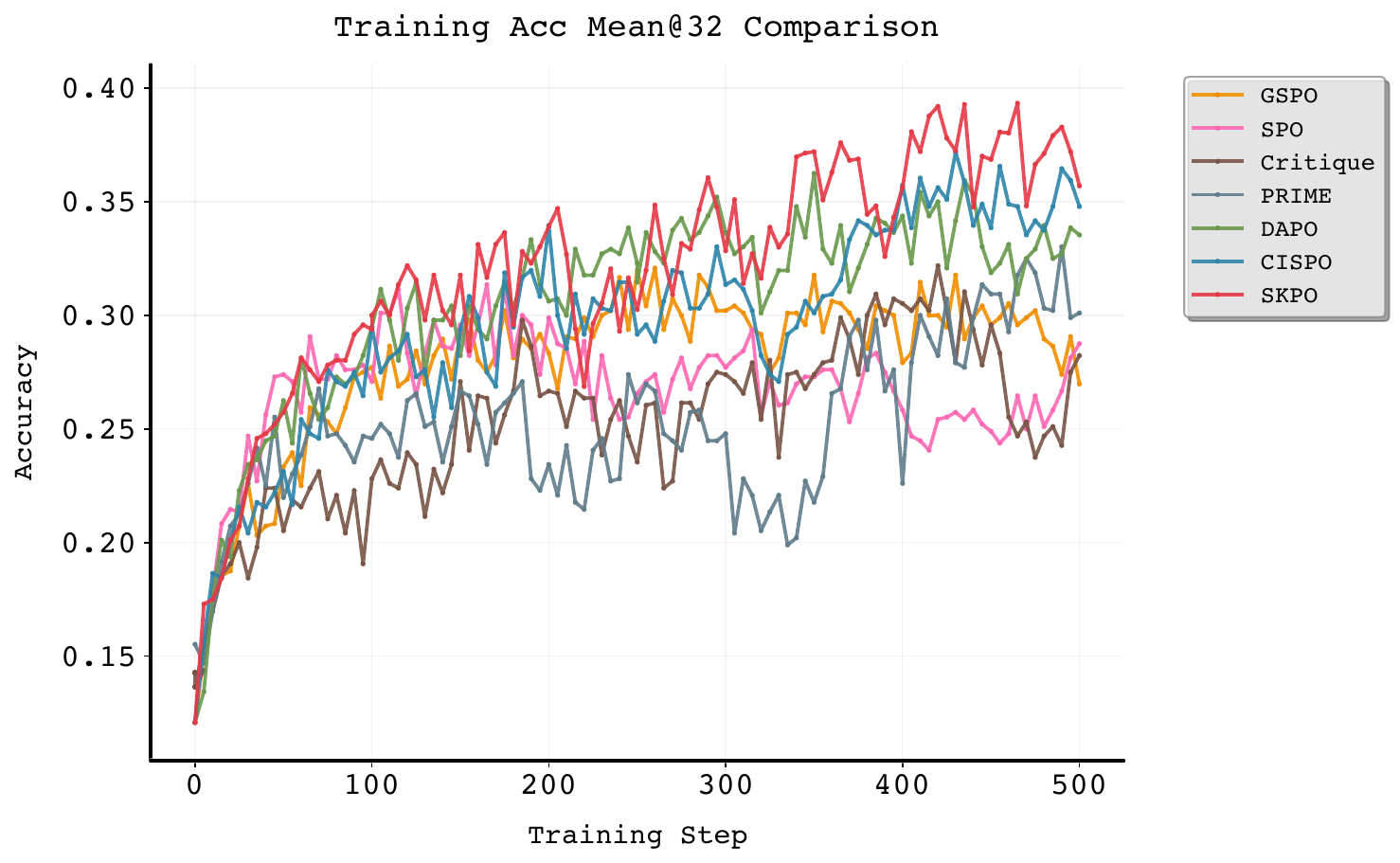}
    \caption{Mean accuracy at 32 samples across training steps.}
    \label{fig:acc_mean_32}
\end{figure}

Figure~\ref{fig:acc_mean_32} shows the mean accuracy (Mean@32) across training, which reflects the average quality across all sampled responses, serving as an indicator of consistent behavior and sample efficiency. \our{} demonstrates superior convergence compared to baseline methods, reaching higher asymptotic performance while maintaining stable training trajectories. Furthermore, \our{} exhibits faster initial improvement, which we attribute to the dense reward signals provided by averaging downstream outcomes for upstream segment optimization; traditional outcome-based methods receive sparse feedback only at trajectory endpoints, while \our{}'s segment-level rewards provide intermediate reward signals that accelerate policy learning in the critical early training phase.

\subsection{Optimization Stability}

\begin{figure}[htbp]
    \centering
    \includegraphics[width=0.85\linewidth]{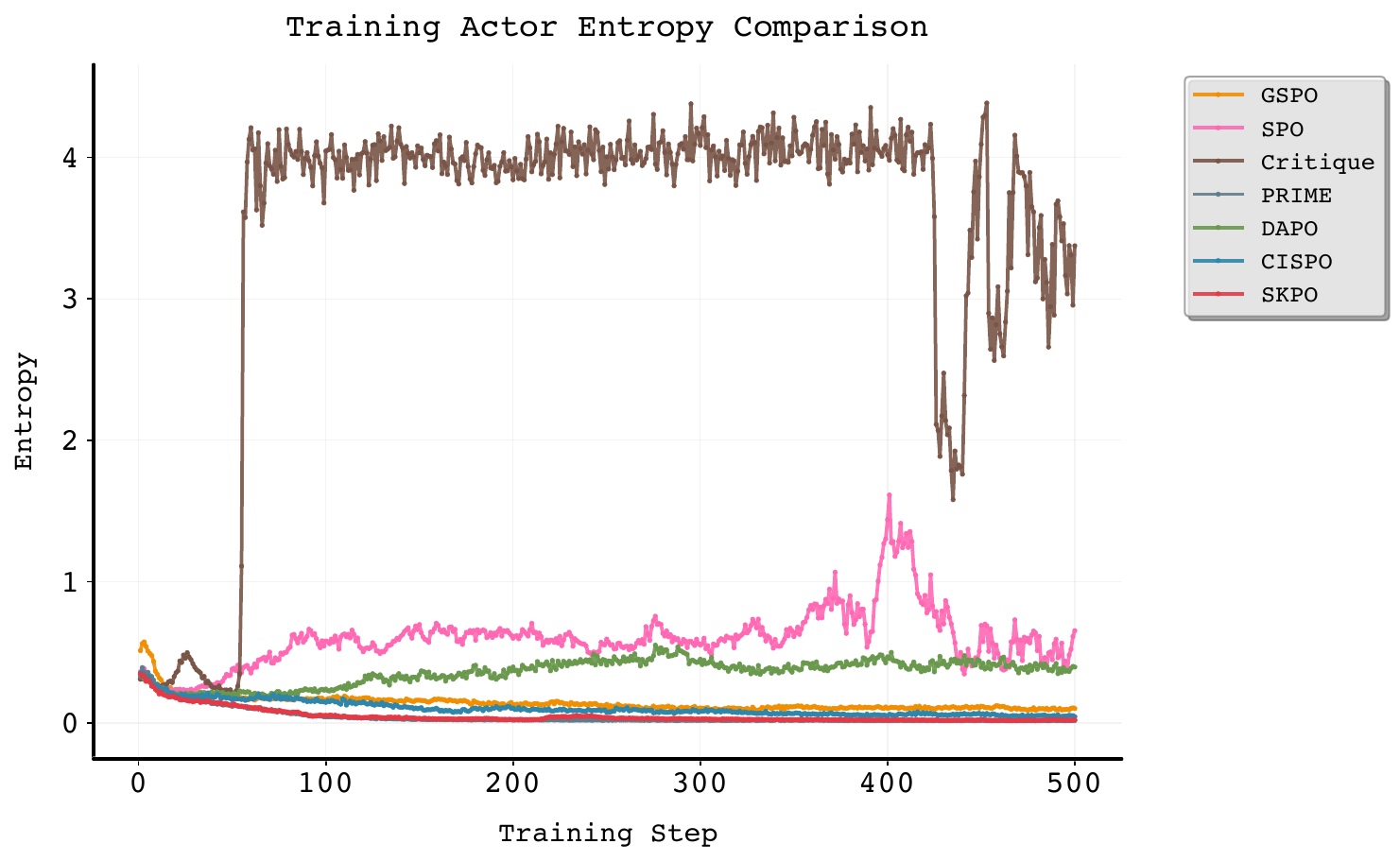}
    \caption{Actor policy entropy throughout training.}
    \label{fig:actor_entropy}
\end{figure}

\paragraph{Entropy Analysis.}
Figure~\ref{fig:actor_entropy} shows the actor policy entropy across training. Most methods exhibit decreasing entropy over the 500-step training schedule, with DAPO being the exception due to its explicit entropy regularization. This entropy reduction is a common behavior as policies converge toward higher-reward solutions. During our exploration of various algorithmic components, we also designed and experimented with some entropy-increasing mechanisms. Unfortunately, we found no clear correlation between entropy trends and final performance, suggesting that our method may not be well-suited for improvements in this direction.

\begin{figure}[htbp]
    \centering
    \includegraphics[width=0.85\linewidth]{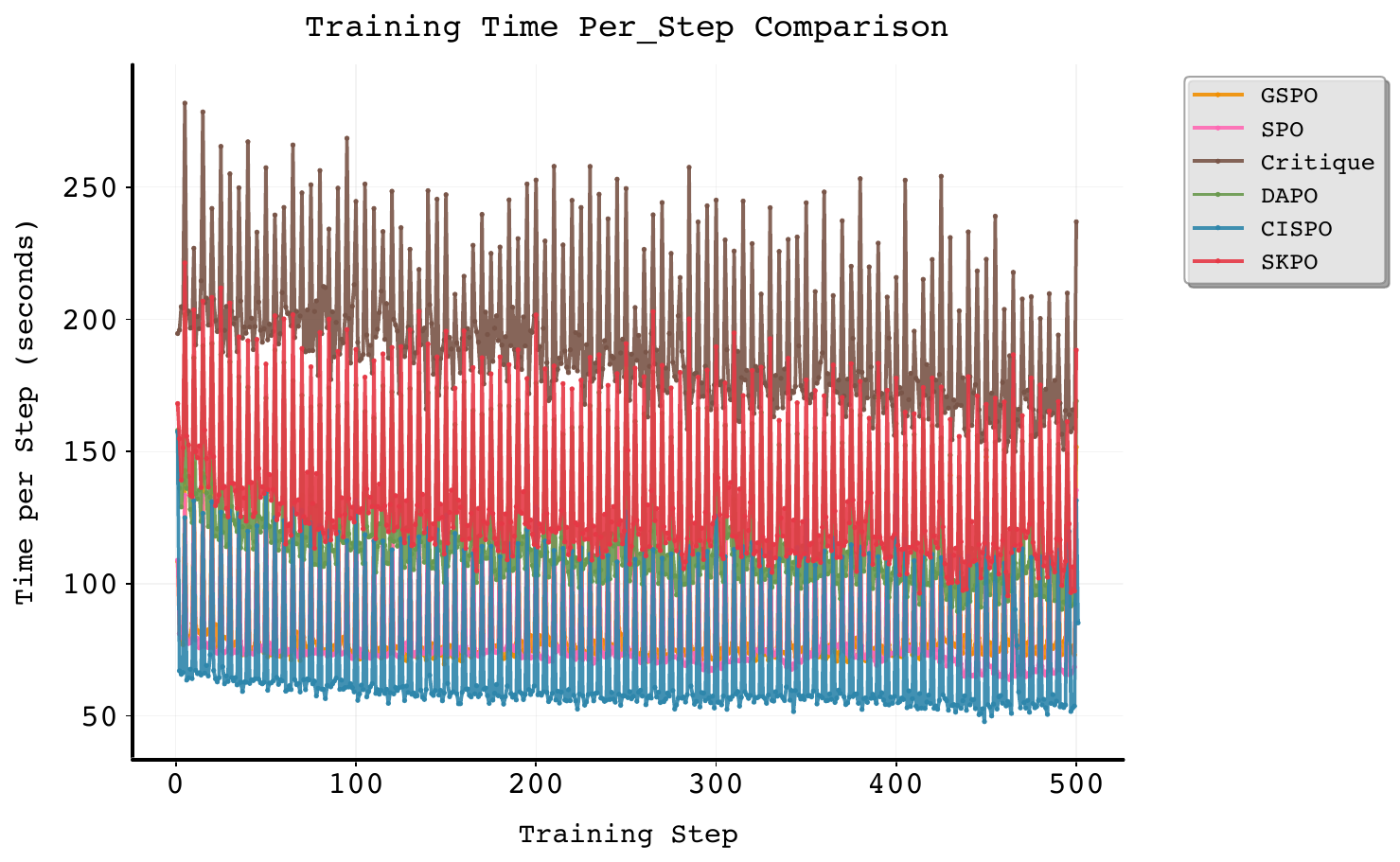}
    \caption{Training time per step across different methods. Critique-GRPO exhibits significantly higher per-step time due to its two-batch rollout strategy: first generating 7 direct responses, then spawning 1 self-critique response in a separate GPU batch. Despite the two-phase upstream-downstream architecture, \our{} achieves comparable wall-clock time to single-phase baselines through our engineering optimization that performs mid-generation KV cache rewriting within a single rollout pass.}
    \label{fig:time_per_step}
\end{figure}

\paragraph{Training Time.}
Figure~\ref{fig:time_per_step} compares the wall-clock training time per optimization step across methods. Critique-GRPO serves as a representative example of two-batch approaches: it first generates 7 direct rollouts, then performs a separate batch dispatch to generate 1 self-critique response conditioned on the best direct response. This sequential processing incurs substantial overhead. A naive two-batch implementation of \our{} would similarly suffer from this overhead. Our engineering optimization eliminates this by modifying the vLLM inference engine to perform mid-generation KV cache rewriting within a single continuous rollout, achieving computational parity with GRPO-family baselines.

\section{Experimental Setup and Hyperparameters}
\label{app:experimental_setup}

This section provides comprehensive details on the experimental configuration, training procedures, and evaluation protocols used throughout this work. All experiments are designed to ensure fair comparison across methods and reproducibility of results.

\paragraph{Dataset.}
We train all models on a curated mathematical reasoning dataset containing 17,398 problems after deduplication (dapo-math-17k). The training data consists of diverse mathematical problems spanning arithmetic, algebra, geometry, number theory, and competition mathematics. We evaluate on standard benchmarks including AIME 2024/2025 (American Invitational Mathematics Examination), AMC (American Mathematics Competition), MATH, Minerva Math, and Olympiad-level problems. For out-of-domain evaluation, we use MMLU-Pro (general reasoning across multiple subjects) and LiveCodeBench (code generation tasks). All training data is formatted with a consistent prompt structure, and responses are truncated to fit within maximum length constraints.

\paragraph{Model Architectures.}
We conduct experiments on two base models: \textbf{Qwen2.5-Math-7B}, a 7-billion parameter language model specialized for mathematical reasoning, and \textbf{Llama-3.2-3B-Instruct}, a 3-billion parameter instruction-tuned model. Both models use the same tokenizer as their respective base model families. For \our{}, the upstream and downstream phases share the same model parameters, with only the input conditioning (original problem $q$ vs. augmented input $[s, q]$) differing between phases.

\paragraph{Optimization Hyperparameters.}
We train all models for \textbf{500 optimization steps} with a \textbf{batch size of 128 prompts} per step. The learning rate is set to $1 \times 10^{-6}$ with linear warmup over the first 10 steps and no decay thereafter. We use the AdamW optimizer with weight decay $0.1$, $\beta_1 = 0.9$, and $\beta_2 = 0.999$. Gradient norms are clipped at $1.0$ to improve training stability. We employ mixed-precision training (bfloat16) to accelerate computation and reduce memory usage. The policy gradient mini-batch size is set to 32 prompts, meaning each optimization step processes 4 mini-batches ($128 / 32 = 4$). Loss aggregation uses token-mean normalization, where gradients from all tokens contribute equally after normalizing by response length.

\paragraph{\our{} Configuration.}
For \our{}, we set the downstream group size to $G=8$ responses per upstream segment, providing sufficient samples for stable Monte Carlo reward estimation. The upstream segment termination position is sampled uniformly from $[\frac{1}{6}L_p, \frac{1}{2}L_p]$, where $L_p$ is the per-prompt average response length maintained using the same SPO-style KL-adaptive update mechanism as the value tracker (initialized to 1024 tokens). This statistical approach adapts to varying problem complexity without relying on formatting conventions; maximum segment length is implicitly bounded by the maximum response length (3072 tokens). The asymmetric optimization strategy employs different configurations for each phase. \textbf{Upstream (SPO)} uses temporal baseline $\hat{V}(q)$ with KL-adaptive forgetting factor, with $\rho_{\min} = 0.875$, $\rho_{\max} = 0.96$, and half-life $\tau_{1/2} = 8.0$ steps. Unlike the original SPO which collects $n_0$ samples before initializing the tracker, we initialize the baseline to zero (neutral in reward space) and update immediately upon the first observation for practical training efficiency; we enable prioritized sampling with $\epsilon = 0.05$ for curriculum learning; batch size is 128 prompts $\times$ 1 response = 128 samples. \textbf{Downstream (GRPO variant)} uses group-relative advantages with importance sampling clipping, with clipping threshold $\varepsilon = 0.2$ for the standard range $[0.8, 1.2]$ and an additional upper bound at $1.28$ (Clip-Higher); batch size is 128 prompts $\times$ 8 responses = 1024 samples. We balance upstream and downstream objectives with equal weights ($w_{\text{up}} = w_{\text{down}} = 0.5$), normalize advantages using batch statistics ($\tilde{A} = (A - \mu_{\mathcal{B}}) / \sigma_{\mathcal{B}}$), and disable KL penalty ($\beta = 0$) and entropy regularization ($\alpha_{\text{ent}} = 0$) to focus purely on outcome-based optimization.

\paragraph{Baseline Configurations.}
All baselines use the same training schedule (500 steps), batch size (128 prompts), learning rate ($1 \times 10^{-6}$), and base model to ensure fair comparison. Since SPO generates only 1 rollout per prompt, its batch size is scaled to 1024 prompts to match the total sample count of other methods with group size 8. Baseline-specific hyperparameters follow their respective original papers.

\section{Prompt for Implicit Advantage Analysis}
\label{app:prompt}

For the implicit advantage analysis in \Cref{sec:advanced_analysis}, we use an external model (GPT-5-nano) to perform Monte Carlo continuation sampling. The prompt used for continuation is shown below:

\begin{tcolorbox}[notitle, sharp corners, breakable, colframe=black, colback=gray!5, 
       boxrule=1.5pt, boxsep=2pt, enhanced]
       \footnotesize
       {\fontfamily{pcr}\selectfont
\begin{lstlisting}
Solve the following problem. Provide detailed reasoning, and you MUST end your response with the final answer wrapped in \boxed{}, for example: \boxed{42}.

Problem:
{problem}

Here is a partial solution to continue from:

{segment}

Please continue from exactly where this solution left off. Do NOT repeat any work already done. Complete the solution and provide the final answer in \boxed{}.
\end{lstlisting}
}
\end{tcolorbox}

\noindent where \texttt{\{problem\}} is the mathematical problem statement, and \texttt{\{segment\}} is the partial reasoning from the trained model at a specific relative position. The external model is instructed to continue from exactly where the partial solution left off, without repeating prior work. By sampling multiple continuations and computing the success rate, we obtain the segment-level reward estimate that reflects ``how easy is it to reach a correct answer from this intermediate state.''

\paragraph{Why Instruction-Based Continuation?} Ideally, one would simply concatenate the problem and partial response as a prefix, then let the model generate completions directly. However, current commercial LLM APIs do not support raw text completion without chat templates, as they require messages formatted according to their instruction-following interface. Therefore, we cannot perform true prefix-based continuation. Instead, we rely on the external model's instruction-following capability by explicitly instructing it to ``continue from exactly where this solution left off'' and ``do NOT repeat any work already done.'' This approach leverages the strong instruction-following abilities of advanced models to approximate the desired continuation behavior.

\end{document}